\documentclass{article}
\usepackage{ijcai25}

\usepackage{times}
\usepackage{soul}
\usepackage{url}
\usepackage[hidelinks]{hyperref}
\usepackage[utf8]{inputenc}
\usepackage[small]{caption}
\usepackage{graphicx}
\usepackage{amsmath}
\usepackage{amsthm}
\usepackage{booktabs}
\usepackage{algorithm}
\usepackage{algorithmic}
\usepackage[switch]{lineno}

\usepackage{subfigure}
\usepackage{amsfonts}
\usepackage{multirow}
\usepackage{makecell}

\newtheorem{definition}{Definition}

\title{Variational Offline Multi-agent Skill Discovery}

\author{
Jiayu Chen$^1$
\and
Tian Lan$^2$\and
Vaneet Aggarwal$^{3}$
\affiliations
$^1$Carnegie Mellon University\\
$^2$The George Washington University\\
$^3$Purdue University\\
\emails
jiayuc2@andrew.cmu.edu,
tlan@gwu.edu,
vaneet@purdue.edu
}

\begin{document}

\maketitle

\begin{abstract}

Skills are effective temporal abstractions established for sequential decision making, which enable efficient hierarchical learning for long-horizon tasks and facilitate multi-task learning through their transferability. Despite extensive research, research gaps remain in multi-agent scenarios, particularly for automatically extracting subgroup coordination patterns in a multi-agent task. In this case, we propose two novel auto-encoder schemes: VO-MASD-3D and VO-MASD-Hier, to simultaneously capture subgroup- and temporal-level abstractions and form multi-agent skills, which firstly solves the aforementioned challenge. An essential algorithm component of these schemes is a dynamic grouping function that can automatically detect latent subgroups based on agent interactions in a task. Further, our method can be applied to offline multi-task data, and the discovered subgroup skills can be transferred across relevant tasks without retraining. Empirical evaluations on StarCraft tasks indicate that our approach significantly outperforms existing hierarchical multi-agent reinforcement learning (MARL) methods. Moreover, skills discovered using our method can effectively reduce the learning difficulty in MARL scenarios with delayed and sparse reward signals. The codebase is available at https://github.com/LucasCJYSDL/VOMASD.

\end{abstract}

\section{Introduction}

Skill discovery aims at extracting useful temporal abstractions from decision-making sequences. The downstream policy learning can be much more efficient by simply composing the discovered skills temporally into complex maneuvers. Also, skills can potentially be transferred among tasks to facilitate multi-task learning. Despite considerable research on single-agent skill discovery \cite{DBLP:conf/iclr/EysenbachGIL19,DBLP:conf/nips/ChenAL23}, skill discovery in multi-agent reinforcement learning (MARL) remains under-explored.
A straightforward approach is to discover single-agent skills for each agent independently and then learning a multi-agent meta policy to coordinate their use, as in \cite{DBLP:conf/atal/YangBZ20,DBLP:conf/gecco/SachdevaKMT21}. However, multi-agent coordination can not be abstracted in such individual skills. 
On the other hand, there are a limited number of works  \cite{DBLP:conf/nips/ChenCLA22,DBLP:conf/nips/Yang0LZL23} on discovering skills for the entire team of agents. However, in multi-agent tasks, coordination patterns can emerge within subgroups of varying scales (from $1$ to $n$), and team skills (i.e., $n$-agent skills) can be inflexible to use.

Notably, complex multi-agent tasks can often be decomposed as a series of subtasks, each of which requires participation of a subgroup of agents for a certain duration. The policies for these subgroups can be abstracted as multi-agent skills.
While agents can explore various forms of collaboration in an online setting (by interacting with the environment), offline multi-agent skill discovery in contrast must infer latent coordination patterns from agent interactions in the offline data, with the subgroup size arbitrarily varying from 1 to $n$. 
This gives rise to a combinatorial problem of dynamic subgroup division and forming temporal abstractions within each subgroup for skill discovery, which is a significant new challenge. \textbf{To the best of our knowledge, this is the first work to fully automate the extraction of collaborative patterns among agents as subgroup skills from offline data.}
We also note that the problem is different from 
 (online) role-based MARL \cite{DBLP:conf/aaai/XuBZ0F23,zhou2024constructing}. They instead focus on partitioning agents into subdivisions that consist of agents with similar responsibilities (i.e., roles), sharing the same policy and thus homogeneous behaviors. Our goal is to learn multi-agent skills -- a collective set of single-agent skills taken by a subgroup where agents could have distinct yet coordinated behaviors.

To be specific, we provide effective auto-encoder frameworks for extracting embeddings of subgroup coordination patterns from offline data as a codebook, where each code corresponds to a multi-agent skill and provides abstractions in both subgroup- and temporal-level. We propose two scheme designs for this purpose: VO-MASD-3D and VO-MASD-Hier. In VO-MASD-3D, three-dimensional codebooks are adopted, where each multi-agent skill code consists of several single-agent skill codes such that it can be used to represent subgroup behaviors. In contrast, VO-MASD-Hier employs a two-level codebook: the bottom codes encode individual behaviors, while each top code is aggregated from a set of bottom codes to encode subgroup behaviors.
Further, to enable automatic grouping while forming temporal abstractions, we co-train a grouping function with the proposed auto-encoder schemes. Using this function, agents can be dynamically grouped based on the environment state, and each subgroup can then be assigned a multi-agent skill of the corresponding size. 
Importantly, our algorithm is designed to work with multi-task data, such that the discovered skills can be utilized in multiple relevant tasks (without retraining). 
Empirical results on challenging StarCraft tasks \cite{DBLP:conf/atal/SamvelyanRWFNRH19,ellis2024smacv2} demonstrate the superiority of the discovered multi-agent skills using our algorithm even in previously unseen tasks, and show the great advantages brought by the use of skills in long-horizon multi-agent tasks characterized by sparse reward signals.

\section{Background} \label{bg}

\textbf{Dec-POMDP:} This work focuses on a fully cooperative multi-agent setting with only partial observation for each agent, which can be modeled as a decentralized partially observable markov decision process (Dec-POMDP) \cite{oliehoek2016concise} and described with a tuple $G = \langle n, I, S, O, F, A, \mu, P, R, \gamma \rangle$. At a time step, each agent $i \in I = \{1, \cdots, n\}$ would obtain a local observation $o^i \in O$ from the observation function $F(s, i): S \times I \rightarrow O$, where $s$ is the real state of the environment, and determine its action $a^i \in A$. This would lead to a state transition in the environment according to the function $P(s'\mid s, \vec{a}): S \times A^n \times S \rightarrow [0, 1]$ and all agents would receive a shared team reward $r = R(s, \vec{a}): S \times A^n \rightarrow \mathbb{R}$. To mitigate the issue of partial observability, each agent $i$ holds an action-observation history $\tau^i_{t-1} = (o_1, a_1, \cdots, o_{t-1}, a_{t-1})$ and decides on its action $a^i_t$ based on a policy $\pi^i(a^i_t\mid o^i_t, \tau^i_{t-1})$. The goal of MARL in a Dec-POMDP can be formally defined as $\max_{\vec{\pi}} \mathbb{E}_{\mu, \vec{\pi}, P, R}\left[\sum_{t=0}^{\infty} \gamma^t r_t\right]$, where $\vec{\pi} = (\pi^1, \cdots, \pi^n)$ and $\mu(s_0): S \rightarrow [0, 1]$ denotes the distribution of the initial state. 


\textbf{CTDE:} The paradigm of centralized training with decentralized execution (CTDE) \cite{DBLP:journals/jair/OliehoekSV08} is proposed for solving Dec-POMDP and has gained substantial attention. In this paradigm, agents learn their policies with access to global information (e.g., the state $s$) during centralized training and only rely on their local action-observation histories for decentralized execution. 
Notably, the discovered skills with our algorithm can be easily integrated into the CTDE paradigm. We select MAPPO \cite{DBLP:conf/nips/YuVVGWBW22} as the base CTDE MARL algorithm throughout this work, as it has shown superior performance across various MARL benchmarks. It learns a decentralized actor $\pi(a^i_t\mid o^i_t, \tau^i_{t-1})$, which is used for each agent $i \in I$ to determine its action $a^i_t$ based on its individual action-observation history $(o^i_t, \tau^i_{t-1})$, and a centralized critic $V(s_t)$. Viewing the $n$ agents as a whole, the critic function is trained as in a single-agent RL algorithm -- PPO \cite{DBLP:journals/corr/SchulmanWDRK17}, while the actor (of each agent) is trained to maximize the advantage function defined with the team reward and centralized critic.

\textbf{Skill \& Task Decomposition:} In single-agent scenarios, skills are used as temporal abstractions of an agent's behaviors. This is inspired by the fact that complex tasks can usually be decomposed as a sequence of subtasks and each subtask can be handled with a corresponding subpolicy, i.e., a skill. With skills, an agent learns a hierarchical policy, where the low-level part $\pi_l(a\mid s, z)$ is the skill policy and the high-level part $\pi_h(z\mid s)$ determines the skill selection. Each skill $z \in \Omega^z$, after being selected, will be executed for $H$ time steps -- a predefined subtask duration. 
However, in multi-agent scenarios, task decomposition occurs not just at the temporal level but also at the agent level, since the overall multi-agent task can be viewed as several subgroup tasks executed in parallel. Here, we define such a task decomposition:
\begin{definition}
  Given a cooperative multi-agent task $\langle n, I, S, O, F, A, \mu, P, R, \gamma \rangle$, at a time step, it can be decomposed into a set of $m$ subtasks, each of which is solved by a subgroup of agents for $H$ time steps and can be represented as a tuple $\langle n_j, I_j, S, O, F, A, \mu, P, R_j, \gamma \rangle$. Here, $\sum_{j=1}^m n_j = n$, $\cup_{j} I_j = I$, and $I_j \cap I_k = \emptyset$ $(\forall\ j \neq k)$.
\end{definition}
Certain subtasks may frequently occur, such as passing and cutting cooperation among two or three players in a football match, and their subpolicies, showing coordination patterns, can be extracted as multi-agent skills and transferred across similar tasks for reuse. 
In this work, we propose an algorithm for discovering such multi-agent skills $Z \in \Omega^Z$ from multi-agent interaction data.



\textbf{Related Works:} \textbf{In Appendix \ref{rw}, we provide a thorough review of research on applying skills in MARL, including MARL with single-agent skills, role-based MARL, and team skill discovery.} We also compare our algorithm with each category of work to highlight our contributions, which we strongly encourage readers to review. As a summary, research on multi-agent skill discovery is still at an early stage of development, especially for the offline setting. Even without prelearned skills, when dealing with a complex multi-agent task, the agents would implicitly learn to decompose the overall task into several subtasks, assign a subgroup for each subtask, and develop a joint policy (i.e., multi-agent skill) within the subgroup to handle the corresponding subtask. Replacing primitive actions with single-agent skills or role policies could make such a learning process more efficient, as agents can assemble these higher-level abstractions to obtain the required subgroup policies more easily. As the first offline multi-agent skill discovery algorithm, our work takes one step further by directly identifying subgroups, which could change throughout a decision horizon, and extracting their coordination patterns as multi-agent skills. With these joint skills, the MARL process could be greatly simplified, since agents only need to select correct multi-agent skills without considering grouping with others or forming subgroup policies. Thus, multi-agent skills represent a more efficient form of knowledge discovery.


\section{Proposed Approach} \label{algo}

\begin{figure*}[t]
\centering
\includegraphics[width=5.3in, height=1.3in]{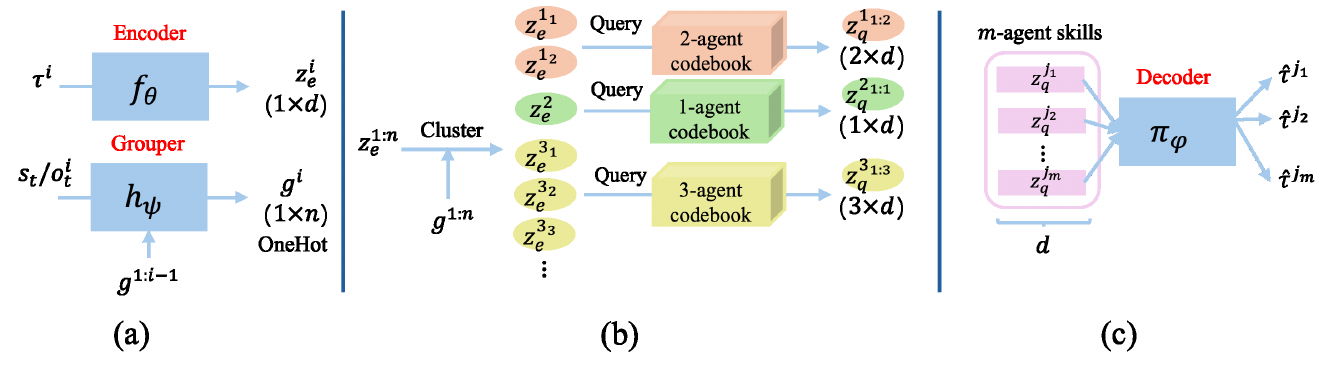}
\vspace{-.1in}
\caption{Multi-agent skill discovery based on a VQ-VAE with 3D codebooks.}
\label{fig:1} 
\vspace{-.05in}
\end{figure*}

Variational Offline Multi-agent Skill Discovery (VO-MASD) aims to extract a finite set of multi-agent skills from given offline trajectories. Proposed for Computer Vision, VQ-VAE \cite{DBLP:conf/nips/OordVK17} provides a fundamental manner to learn discrete representations for complex, high-dimensional data. Besides the encoder and decoder as used in VAEs \cite{DBLP:journals/corr/KingmaW13}, a codebook containing a finite set of codes, each of which is a latent representation of the data, is learned. In this case, VQ-VAE is a natural choice for skill discovery, with each code working as a skill embedding $Z$. Each $Z$ would correspond to a skill policy $\pi_l(\vec{a} \mid \vec{\tau}, Z)$ that leads to continuous multi-agent behaviors. In this section, we present two schemes of VO-MASD based on VQ-VAE by adopting novel codebook designs and involving an automatic grouping module. The challenge is to extract temporal-level abstractions (i.e., useful control sequences) and agent-level abstractions (i.e., multi-agent coordination) at the same time, \textbf{without using domain knowledge or task-specific reward signals}. In this way, VO-MASD can be applied to a mixture of multi-task data and the learned skills are generalizable to a distribution of relevant tasks.




\subsection{VO-MASD based on 3D Codebooks} \label{VO-MASD-3D}

VQ-VAE typically adopts a 2D codebook $[e_1, \cdots, e_k] \in \mathbb{R}^{k \times d}$, where $e_i \in \mathbb{R}^{1 \times d}$ is a latent representation. However, in our case, there are three levels of abstractions: primitive actions $\rightarrow$ single-agent skills $\rightarrow$ multi-agent skills. As part of our novelty, we propose to use 3D codebooks within $\mathbb{R}^{k \times m \times d}$ to represent a set of (i.e., $k$) $m$-agent skills. Each code $e_i = [e_{i, 1}, \cdots, e_{i, m}] \in \mathbb{R}^{m \times d}$ represents a multi-agent skill composed of $m$ single-agent skills. Note that $m$ ranges from $1$ to $n$ (i.e., the size of the team).

A straightforward approach to utilize such codebook design for skill discovery is repeatedly applying a VQ-VAE with an $m$-agent codebook to $m$-agent skill discovery, for $m=1, \cdots, n$. 
As some variational methods for single-agent skill discovery \cite{DBLP:conf/icml/CamposTXSGT20,DBLP:conf/iclr/AjayKALN21}, the objective for learning $m$-agent skills could be minimizing the reconstruction error of $m$-agent trajectory segments. Ideally, after training, each code can represent a coordination pattern among $m$ agents and the code-conditioned decoder can be used as an $m$-agent skill policy. However, if there are no coordination involving $m$ agents in the offline data, the effort to discover $m$-agent skills would be wasted. Also, in this way, the learning processes for skills involving different numbers of agents are independent and cannot benefit from each other.


In this case, we introduce a grouping function $h_\psi$ that dynamically groups agents throughout an episode to identify existing coordination patterns in the offline data and unify the training of skills with different numbers of agents.
The skill discovery process is illustrated as Figure \ref{fig:1}. 
As shown in (a), at time step $t$, for each agent $i$, we encode its following $H$ time steps, i.e., $\tau^i = [o^i_t, a^i_t, \cdots, o^i_{t+H-1}, a^i_{t+H-1}]$, into a skill embeddings $z^i_e$ using the encoder $f_\theta$. Also, each agent $i$ selects its group based on the global state $s_t$ and group choices of previous agents $g^{1:i-1}$ using a grouping function $h_\psi$. There can be at most $n$ groups, when all agents choose to use individual skills. Notably, both $h_\psi$ and $f_\theta$ are shared by all agents.
Subsequently, in (b), the skill embeddings $z^{1:n}_e$ from the encoder are first clustered based on the grouping result $g^{1:n}$: if \( m \) agents choose the same group (indicated by the one-hot output $g^i$), they aim to form an $m$-agent coordination skill and their respective embeddings will be concatenated in the sequence of their agent indices, resulting in an \( m \times d \) joint embedding $z^{j_{1:m}}_e$. Then, as in VQ-VAE, the code that is the closest to $z^{j_{1:m}}_e$ in the $m$-agent codebook (i.e., $z^{j_{1:m}}_q$) is queried to work as the skill code. 
Finally, in (c), a decoder $\pi_\phi$ maps the skill code back to an $m$-agent trajectory segment, i.e., $\hat{\tau}^{j_{1:m}}$. Taking the subgroup $j_{1:m}$ as an example, the training objective is:
{\small \begin{equation} \label{masd-3d}
\begin{aligned}
     & L^{\text{3D}}(\tau^{j_{1:m}}) =  -\sum_{l=0}^{H-1} \sum_{i=1}^m \log \pi_{\phi}(a^{j_{i}}_{t+l}\mid o^{j_{i}}_{t+l}, z^{j_{i}}_q) \\ &+ \sum_{i=1}^m \left[\|\text{sg}(z^{j_{i}}_e) - e^{j_i}\|_2^2 + \beta \|z^{j_{i}}_e - \text{sg}(e^{j_i})\|_2^2\right] 
\end{aligned}
\end{equation}}

\begin{figure*}[t]
    \centering
    \includegraphics[width=5.7in, height=1.1in]{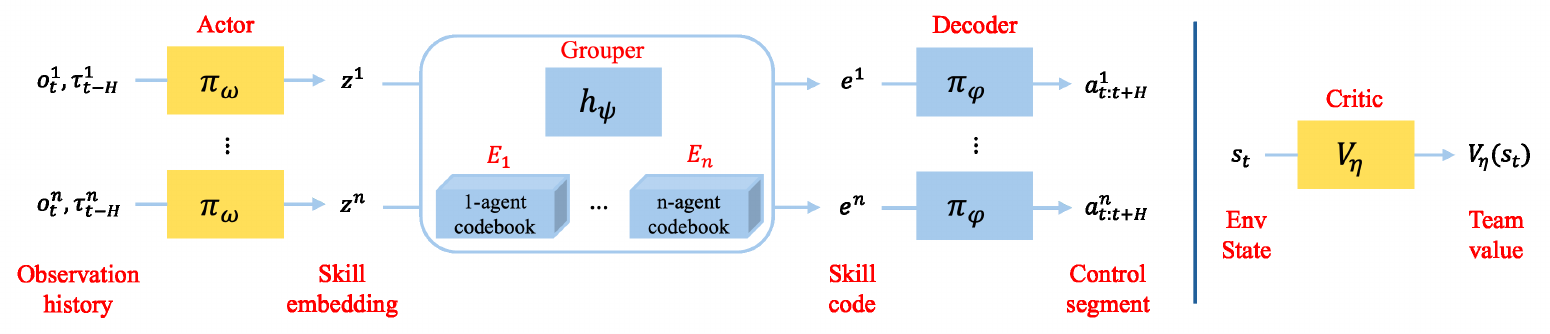}
    \vspace{-.05in}
    \caption{Utilizing discovered skills for downstream CTDE MARL. Compared to standard CTDE MARL, individual actions $a^{1:n}$ are replaced with skill embeddings $z^{1:n}$ as the actor's output. These embeddings are then translated into skill codes and control segments using the pretrained VO-MASD components, as shown in Figure \ref{fig:1}. Thus, only the individual actor $\pi_\omega$ and centralized critic $V_\eta$ need to be trained.}
    \label{fig:ctde}
    \vspace{-.05in}
\end{figure*}

As shown in Figure \ref{fig:1}, $z^{j_i}_e = f_\theta(\tau^{j_i})$, $e^{j_{1:m}} = \arg \min_{e \in E_m} \|z^{j_{1:m}}_e - e\|_2$ ($E_m$ denotes the $m$-agent codebook), and $z^{j_i}_q = e^{j_i}$. $L^{\text{3D}}(\tau^{j_{1:m}})$ is an objective with respect to (w.r.t.) $\theta, \phi, E_m$. 
As in VQ-VAE, the first term in Eq \eqref{masd-3d} is a reconstruction loss of trajectory segments,
and the last two terms move the skill codes (e.g., $e^{j_i}$) and encoder embeddings (e.g., $z^{j_i}_e$) towards each other, where $\text{sg}$ represents the stop gradient operator. Through reconstructing $m$-agent ($m \in \{1, \cdots, n\}$) trajectory segments in an auto-encoder framework, representations of $m$-agent skills can be extracted as codes in the codebook.
The overall objective for VO-MASD-3D is as below:
{\small\begin{equation} \label{real_vomasd_3d}
    \min_{\theta, \phi, E_{1:n}} L^{\text{3D}} =  \min_{\theta, \phi, E_{1:n}} \mathbb{E}_{\tau^{1:n} \sim \mathcal{D}_{H}} \sum_j L^{\text{3D}}(\tau^{j_{1:m}})
\end{equation}}
Here, $\mathcal{D}_{H}$ is a (multi-task) offline dataset with trajectories segmented every $H$ time steps; each $n$-agent trajectory segment is partitioned into subgroups (e.g., $j_{1:m}$) based on the grouping function $h_\psi$. Note that, unlike $f_\theta$, $E_{1:n}$, and $\pi_\phi$, $h_\psi$ cannot be trained in an end-to-end manner by minimizing Eq (\ref{real_vomasd_3d}), since its output $g^{1:n}$ are used for clustering which is not an differentiable operation. Thus, we choose to optimize $h_\psi$ with MAPPO, where each agent $i$ takes an action $g^i$ to maximize the global return $- L^{\text{3D}}$. 
In this way, all modules in Figure \ref{fig:1} are effectively updated with a common objective.

This framework offers several advantages: (1) the training of skills with different number of agents can facilitate each other, as they share all modules but the codebook; (2) the modeling of temporal- and agent-level abstractions within multi-agent skills are decoupled as training the decoder to reconstruct single-agent trajectories and training the grouper for automatic grouping, respectively; (3) the grouper is trained to form subgroups only when it enhances the overall pattern extraction objective (i.e., Eq (\ref{real_vomasd_3d})), ensuring that each subgroup, along with its associated policy, corresponds to a genuine coordination pattern.

\begin{algorithm}[t] 
\caption{MAPPO with learned skills}
\begin{algorithmic} \label{alg:1}
  \STATE Input: $\pi_\omega$, $V_\eta$, $\pi_{\phi}$, $h_\psi$, $E_{1:n}$, $\text{Env}$
  \STATE Initialize $\pi_\omega$, $V_\eta$
  \WHILE{not converged}
  \STATE $\text{Buffer} \leftarrow \emptyset$
  \FOR{$b=1 \cdots B$}
  \STATE Initialize $\tau^{1:n}_{-H}$, $\text{Traj} \leftarrow \emptyset$, $\tilde{r}\leftarrow 0$
  \FOR{$t=0 \cdots T$}
  \IF{$t \% H == 0$}
  \STATE $z^i_t, \tau^i_t \leftarrow \pi_\omega(o_t^i, \tau^i_{t-H})$, $i=1 \cdots n$
  \STATE Get $e^{1:n}$ based on $z^{1:n}_t$ using $h_\psi$ and $E_{1:n}$, following Fig \ref{fig:1} (b)
  \STATE Add $(\tilde{r}, s_t, o_t^{1:n}, \tau^{1:n}_{t-H}, z^{1:n}_t)$ to $\text{Traj}$
  \STATE $\tilde{r} \leftarrow 0$
  \ENDIF
  \STATE $a^i_t \leftarrow \pi_{\phi}(o_t^i|e^i)$, $i=1 \cdots n$
  \STATE $r_t, s_{t+1}, o^{1:n}_{t+1} \leftarrow \text{Env}(a^{1:n}_t)$, $\tilde{r} \mathrel{+}= r_t$
  \ENDFOR
  \STATE $\text{Buffer} \leftarrow \text{Buffer} \cup \text{Traj}$
  \ENDFOR
  \STATE Train $\pi_\omega$, $V_\eta$ based on $\text{Buffer}$ using MAPPO
  \ENDWHILE
\end{algorithmic}
\end{algorithm}

We illustrate how to utilize the discovered skills in CTDE MARL in Figure \ref{fig:ctde}. Also, Alg \ref{alg:1} outlines the detailed training process for a decentralized actor $\pi_\omega$ and centralized critic $V_\eta$ using MAPPO in a multi-agent task $\text{Env}$, leveraging the pre-trained components $h_\psi$, $E_{1:n}$, and $\pi_{\phi}$. In particular, every $H$ time steps, the actor produces a skill embedding $z^i \in \mathbb{R}^{1 \times d}$ for each agent $i$. $z^{1:n}$ are mapped to the closest multi-agent skill codes $e^{1:n}$ using the grouper $h_\psi$ and codebook $E_{1:n}$, following Figure \ref{fig:1} (b). Then, for the next $H$ time steps, each agent $i$ interacts with $\text{Env}$ using corresponding $\pi_\phi(a^i \mid s^i,e^i)$, i.e., the decoder working as the skill policy.
Based on the interaction transitions, i.e., $\{(s_{t}, o_{t}^{1:n}, \tau^{1:n}_{t-H}, z^{1:n}, \tilde{r}_t, s_{t+H})\}$, $\pi_\omega$ and $V_\eta$ can be trained with MAPPO, where $\tau^{1:n}_{t-H}$ are the skill -- observation (i.e., $z-o$) history, $z^{1:n}$ can be viewed as (high-level) actions, and $\tilde{r}_t = \sum_{l=t}^{t+H-1} r_l$ is the skill reward. 

\begin{figure*}[t]
\centering
\includegraphics[width=5.3in, height=1.3in]{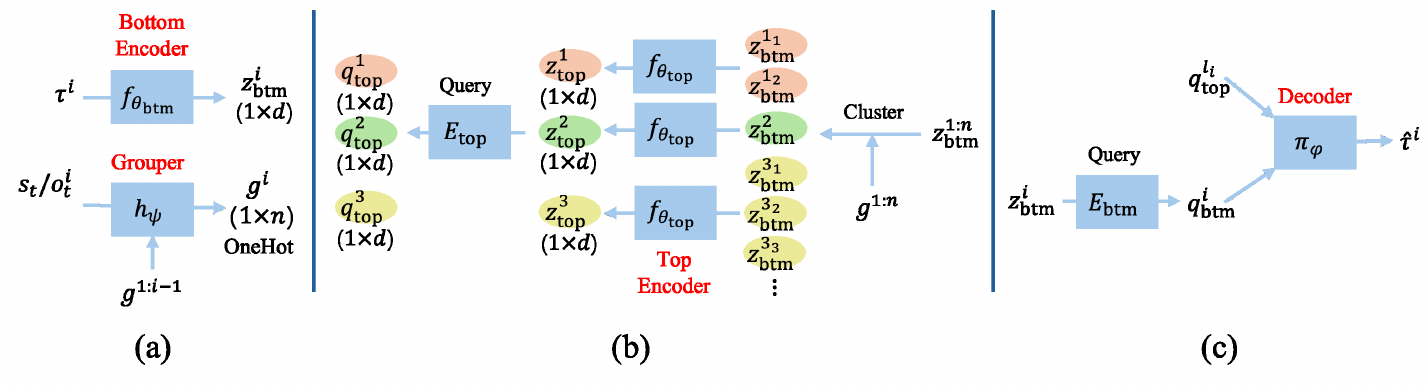}
\vspace{-.1in}
\caption{Multi-agent skill discovery based on a VQ-VAE with a hierarchical codebook.}
\label{fig:2} 
\vspace{-.05in}
\end{figure*}

\textbf{Besides the approach illustrated in Figure \ref{fig:ctde}, we propose two alternative methods for mapping \( z^{1:n} \) to \( e^{1:n} \) in Appendix \ref{GAMSEC}.} All the three methods assign each multi-agent ($m \times d$) code as a complete unit to a corresponding-size ($m$-agent) subgroup, such that the collaboration pattern encoded in the multi-agent code can be utilized. Alternatively, each ($m \times d$) code can be decomposed into a set of ($m$) single-agent codes. Each agent $i$ could then independently select its skill from the set of single-agent codes. In particular, the single-agent skill code closest to the agent's actor output $z^i$ would be selected. We denote this algorithm as `VO-MASD-Mixed'. Later, we empirically compare these four skill assignment manners. Importantly, no matter which manner we choose, we only need to train a decentralized actor and a centralized critic for (online) MARL, without any additional learning effort beyond standard CTDE MARL approaches. 

\subsection{VO-MASD based on a Hierarchical Codebook}

Here, we present VO-MASD-Hier, which is an alternative to VO-MASD-3D and adopts a hierarchical codebook as in \cite{DBLP:conf/nips/RazaviOV19}. Although \cite{DBLP:conf/nips/RazaviOV19} is originally proposed for image generation, its top and bottom codebooks perfectly echo the two-level structure of multi-agent and single-agent skill embeddings.
Thus, we propose to learn top and bottom codebooks as agent- and temporal-level abstractions, respectively, for multi-agent skill discovery. The overall framework of VO-MASD-Hier is shown as Figure \ref{fig:2}. It contains a two-level codebook, i.e., $E_{\text{top}}, E_{\text{btm}}$.
VO-MASD-Hier does not need to learn $n$ codebooks (i.e., $E_{1:n}$) as in VO-MASD-3D, while VO-MASD-3D can potentially make better use of domain knowledge. For example, if the scale of coordination subgroups (e.g., $m$) is known in advance, VO-MASD-3D only needs to learn $E_1$ and $E_m$, while VO-MASD-Hier can not specify the number of agents within a multi-agent skill. 


The skill discovery process is detailed as follows. In Figure \ref{fig:2} (a), the embedding process of each individual trajectory segment $\tau^i$ ($i=1, \cdots, n$) is the same as the one of VO-MASD-3D (i.e., Figure \ref{fig:1} (a)). Subsequently, in (b), the skill embeddings $z^{1:n}_{\text{btm}}$ are clustered based on the output from the grouping function, i.e., $g^{1:n}$, and then embeddings within the same subgroup (e.g., $z_{\text{btm}}^{1_{1:2}}$) are aggregated to a higher-level representation (e.g., $z_{\text{top}}^{1}$) which is then used to query a top code (e.g., $q_{\text{top}}^{1}$). Note that the aggregator $f_{\theta_{\text{top}}}$ uses a multi-head attention module \cite{DBLP:conf/nips/VaswaniSPUJGKP17} to process varied-length inputs and so can be shared by all subgroups (of varied sizes). 
In Figure \ref{fig:2} (c), for each agent $i$, a bottom code $q_{\text{btm}}^i$ is assigned based on its individual skill embedding $z^{i}_{\text{btm}}$. Finally, $q_{\text{btm}}^i$ and $q_{\text{top}}^{l_i}$, involving temporal- and agent-level abstractions respectively, are used to decode/reconstruct $\tau^i$. The overall objective is $ \min_{\theta_{\text{top}, \text{btm}}, E_{\text{top}, \text{btm}}, \pi_{\phi}} \mathbb{E}_{\tau^{1:n} \sim \mathcal{D}_{H}} L^{\text{Hier}}(\tau^{1:n})$:
{\small \begin{equation} \label{masd-hier}
\begin{aligned}
     & L^{\text{Hier}}(\tau^{1:n}) = -\sum_{j=0}^{H-1} \sum_{i=1}^n \log \pi_{\phi}(a^i_{t+j}\mid o^i_{t+j}, q_{\text{btm}}^i, q_{\text{top}}^{l_i}) \\
     & + \sum_{i=1}^n \left[\|\text{sg}(z^{i}_{\text{btm}}) - q^{i}_{\text{btm}}\|_2^2  + \beta \|z^{i}_{\text{btm}} - \text{sg}(q^{i}_{\text{btm}})\|_2^2\right] \\
     & + \sum_{i=1}^n \left[ \|\text{sg}(z^{l_{i}}_{\text{top}}) - q^{l_{i}}_{\text{top}}\|_2^2 + \beta \|z^{l_{i}}_{\text{top}} - \text{sg}(q^{l_{i}}_{\text{top}})\|_2^2\right]
\end{aligned}
\end{equation}
}
This loss function is similar with Eq (\ref{masd-3d}), i.e., to reconstruct the input multi-agent trajectory segment, and move the codes and corresponding skill embeddings towards each other. 

As for the design intuition, considering the first term in Eq (\ref{masd-hier}), the gradient w.r.t. the bottom code $q^{i}_{\text{btm}}$ only comes from reconstructing agent $i$'s individual skill trajectory $\tau^i$. However, the gradient w.r.t. the top code $q^{l_i}_{\text{top}}$ is derived from reconstructing the multi-agent skill trajectories of the subgroup $l_i$ that $i$ belongs to, since each agent $j$ in $l_i$ would adopt $q^{l_i}_{\text{top}}$ as the decoder condition to reconstruct corresponding $\tau^j$. This reflects that the top and bottom codebooks are trained to embed subgroup- and temporal-level abstractions, respectively. 
Notably, both VO-MASD-3D and VO-MASD-Hier follow the inductive bias: primitive actions $\rightarrow$ single-agent skills $\rightarrow$ multi-agent skills. That is, each single-agent skill code is trained to embed an individual trajectory and each multi-agent skill code is a composition of single-agent ones. To be specific, in VO-MASD-3D, each $(m \times d)$ multi-agent code contains a set $(m)$ of $(1 \times d)$ single-agent codes; while for VO-MASD-Hier, each multi-agent embedding $z_{\text{top}}$ is obtained through aggregating individual skill embeddings $z_{\text{btm}}$ from the same subgroup using an attention mechanism, as shown in Figure \ref{fig:2} (b).



To utilize the discovered skills in downstream online MARL, Alg \ref{alg:1} can be applied to VO-MASD-Hier by replacing the process in Figure \ref{fig:1} (b)(c) with corresponding ones in Figure \ref{fig:2} (b)(c). Specifically, a decentralized actor $\pi_\omega$ gives out skill embeddings $z_{\text{btm}}^{1:n}$ every $H$ time steps. $h_\psi$, $E_{\text{top}, \text{btm}}$, $f_{\theta_{\text{top}}}$, and $\pi_\phi$ are fixed during online MARL, transforming $z_{\text{btm}}^{1:n}$ to multi-agent and single-agent skill codes, i.e., $q_{\text{top}}^{1:n}$ and $q_{\text{btm}}^{1:n}$. The decoder is then used to produce skill trajectories of length $H$, according to $\pi_{\phi}(a^i_t\mid o^i_t, q_{\text{top}}^{l_i}, q^{i}_{\text{btm}})$.

%

\section{Evaluation and Main Results} \label{eval}

Experiments are conducted on the StarCraft multi-agent challenge (SMAC) \cite{DBLP:conf/atal/SamvelyanRWFNRH19} -- a commonly-used benchmark for cooperative MARL. Following ODIS \cite{DBLP:conf/iclr/ZhangJLY0Z23}, we adopt two SMAC task sets to test the discovered multi-task multi-agent skills. In each task set, agents control some units like marines, medivacs, and marauders, but the number of controllable agents or enemies varies across tasks in a task set. We refer to the two task sets as `marine' and `MMMs', which evaluate algorithm performance in scenarios with homogeneous and heterogeneous agents, respectively, and are detailed in Appendix \ref{SMACT}. For each task set, we discover skills from offline trajectories of source tasks, and then apply these skills to each task in the task set (including source and unseen tasks) for online MARL. The offline trajectories are collected with well-trained MAPPO agents, which can be viewed as expert demonstrations \textbf{for the source tasks}, and are included in our released code folder. 

Next, we show evaluation results on several aspects. (1) In Section \ref{utility}, we compare skills discovered using different algorithms on the two SMAC task sets, based on their utility for downstream online MARL, to demonstrate the superiority of the multi-agent skills discovered by our methods. (2) In Section \ref{SMACV2}, we test the algorithms on a task set from another benchmark -- SMACv2 \cite{ellis2024smacv2}, which features stochastic environments. (3) In Section \ref{srwd}, we show that, for MARL tasks with sparse reward signals, hierarchical learning with skills discovered using our methods can significantly outperform usual MARL algorithms. Notably, the skills are from relevant but different tasks. (4) \textbf{In Appendix \ref{QADS}}, we visualize the discovered multi-agent skills to evaluate whether they capture meaningful coordination patterns. (5) \textbf{In Appendix \ref{IODQ}}, we test our algorithms on offline datasets of varying quality. (6) \textbf{In Appendix \ref{abl}}, we provide ablation study to show how the components of our algorithm design, including the four skill assignment manners introduced in Section \ref{VO-MASD-3D}, affect the learning performance.

\subsection{Utility of Discovered Skills for Online MARL} \label{utility}

\begin{figure*}[t]
\centering
    \subfigure[3m]{
    \includegraphics[width=2.0in, height=1.15in]{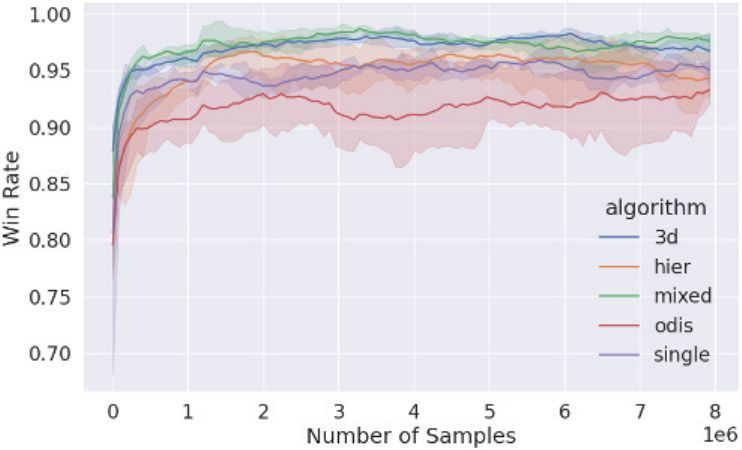}
    \label{2:fig:3(a)}}
    \subfigure[5m]{
    \includegraphics[width=2.0in, height=1.15in]{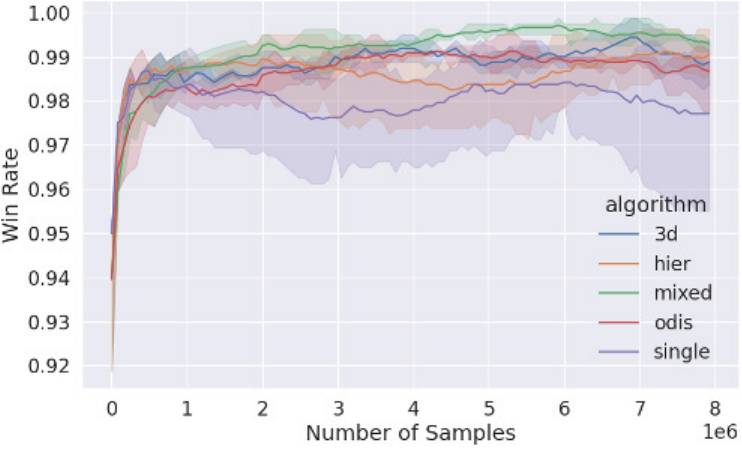}
    \label{2:fig:3(b)}}
    \subfigure[7m]{
    \includegraphics[width=2.0in, height=1.15in]{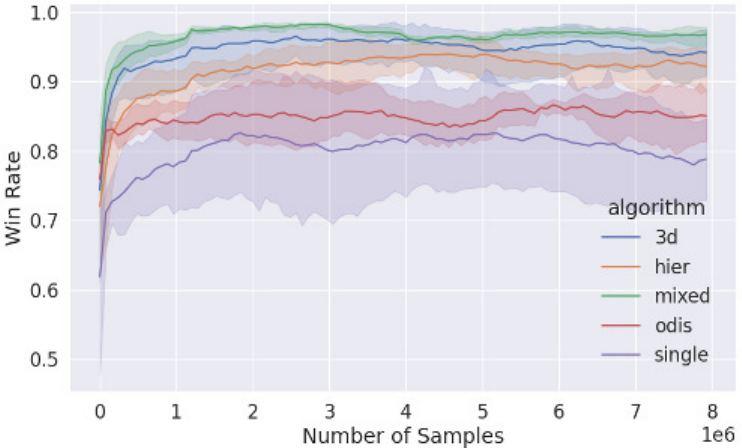}
    \label{2:fig:3(c)}}
    \subfigure[10m]{
    \includegraphics[width=2.0in, height=1.15in]{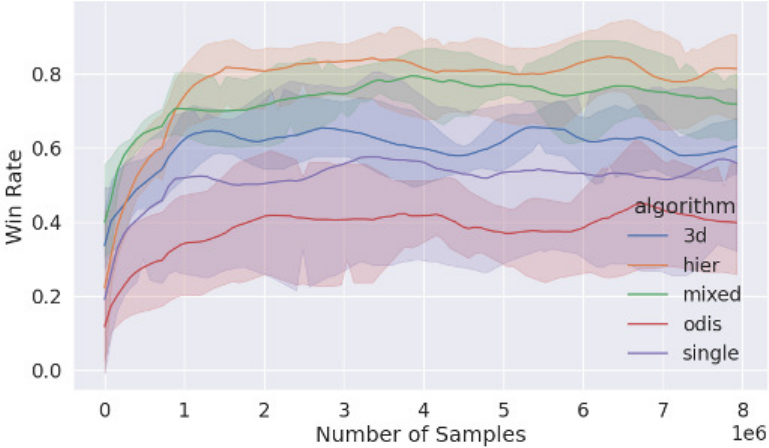}
    \label{2:fig:3(d)}}
    \subfigure[MMM]{
    \includegraphics[width=2.0in, height=1.15in]{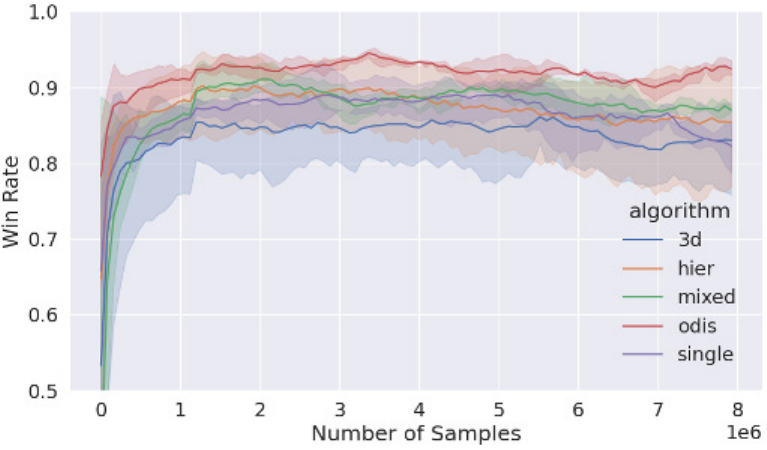}
    \label{2:fig:3(e)}}
    \subfigure[MMM2]{
    \includegraphics[width=2.0in, height=1.15in]{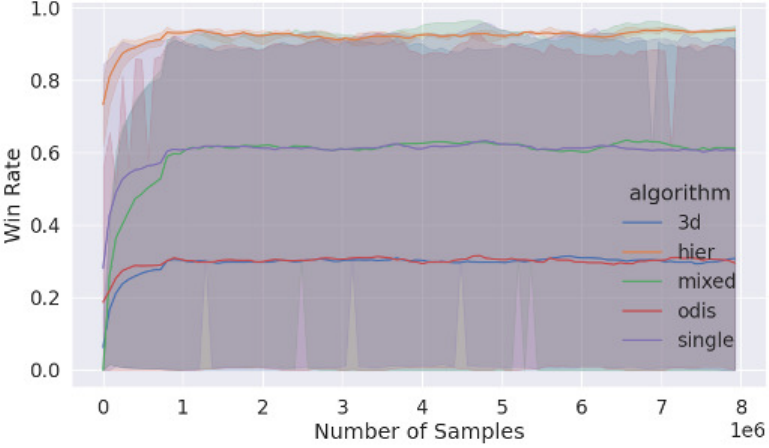}
    \label{2:fig:3(f)}}
\vspace{-.1in}
\caption{Evaluation of effectiveness of the discovered skills using different algorithms for online MARL.}
\label{2:fig:3} 
\vspace{-.1in}
\end{figure*}

The first group of results is shown as Figure \ref{2:fig:3}, where `3d', `hier', `mixed', `single', and `odis' refer to VO-MASD-3D, VO-MASD-Hier, VO-MASD-Mixed, VO-MASD-Single, and ODIS, respectively. As mentioned in Appendix \ref{rw}, ODIS is the only existing algorithm for discovering multi-agent temporal abstractions from offline multi-task data\footnote{In ODIS, the discovered skills are used for offline MARL. For fair comparisons, we instead integrate skills from ODIS with online MARL, as in VO-MASD-3D and VO-MASD-Hier.}, and is a representative of role-based MARL. Notably, ODIS has demonstrated superior performance compared to direct imitation learning from the offline dataset, MADT \cite{DBLP:journals/corr/abs-2112-02845} (an offline MARL algorithm using pretraining), and UPDeT \cite{DBLP:journals/corr/abs-2101-08001} (a SOTA multi-task MARL method), making it a strong baseline for comparison. VO-MASD-Single represents the other main branch of hierarchical MARL -- learning a set of single-agent skills and collaboratively utilizing them for MARL, which is realized through removing $E_{\text{top}}$ and $f_{\theta_{\text{top}}}$ in VO-MASD-Hier (i.e., Figure \ref{fig:2}). VO-MASD-Single discovers and utilizes single-agent skills, while VO-MASD-Mixed discovers multi-agent skills as in VO-MASD-3D but employs the learned skills as single-agent ones, which is detailed in the last paragraph of Section \ref{VO-MASD-3D}. Thus, the baselines include SOTA algorithms in this field and two variations of our algorithms to respectively show the effect of \textbf{discovering} and \textbf{utilizing} skills as multi-agent units. 

Skills (of length 5) discovered from source tasks are applied to both source and unseen tasks for online MARL using Alg \ref{alg:1}. In marine, 3m and 5m are source tasks; while in MMMs, MMM is the source task. \textbf{We believe that the learning performance on unseen tasks with higher-complexity is the best way to testify the utility and generality of skills discovered with different algorithms.} In particular, we track the change of win rates as the number of training samples increases, presenting the mean and 95\% confidence intervals as solid lines and shaded areas, respectively. Several conclusions can be drawn from Figure \ref{2:fig:3}. \textbf{(1)} ODIS and VO-MASD-Single, which represent two main existing approaches of applying skills in MARL, exhibit inferior performance compared to the others, especially in unseen tasks. This underscores the importance of \textbf{discovering} coordination patterns as multi-agent skills, which can significantly enhance performance and generality in new multi-agent tasks. \textbf{(2)} In marine tasks, the performances of VO-MASD-3D and VO-MASD-Hier are comparable, with VO-MASD-Hier performing better in 10m and VO-MASD-3D excelling in the others. However, VO-MASD-3D's performance deteriorates in MMMs, suggesting that its design may not be well-suited for heterogeneous-agent tasks like MMM and MMM2 and indicating a potential future research direction for improvement.
\textbf{(3)} VO-MASD-Mixed follows the same skill discovery process as VO-MASD-3D but adopts the skills as single-agent ones. Surprisingly, VO-MASD-Mixed consistently outperforms VO-MASD-3D. While VO-MASD-3D utilizes fixed combinations of single-agent $(1 \times d)$ codes from the discovery stage, VO-MASD-Mixed explores all possible combinations of these $(1 \times d)$ codes to achieve a higher return, which explains its better performance. However, in the most challenging settings (i.e., 10m and MMM2), VO-MASD-Hier demonstrates better results, showing the potential benefit of \textbf{utilizing} discovered multi-agent skills as complete units. 
\textbf{(4)} The evaluation on MMM2 -- a super-hard task setting \cite{DBLP:conf/atal/SamvelyanRWFNRH19}, demonstrates the superiority of VO-MASD-Hier over other algorithms. All algorithms, except for VO-MASD-Hier, exhibit large variance across different runs.


\subsection{Evaluation Results on SMACv2} \label{SMACV2}

\begin{figure*}[t]
\centering
    \subfigure[Terran-3]{
    \includegraphics[width=2.0in, height=1.15in]{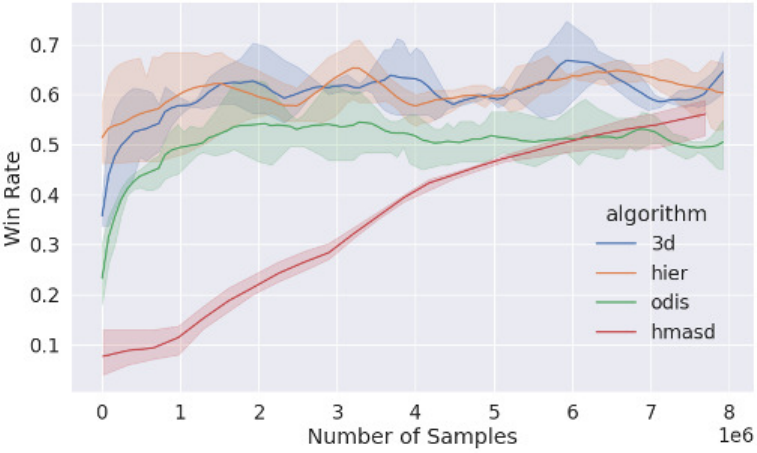}
    \label{fig:8(a)}}
    \subfigure[Terran-5]{
    \includegraphics[width=2.0in, height=1.15in]{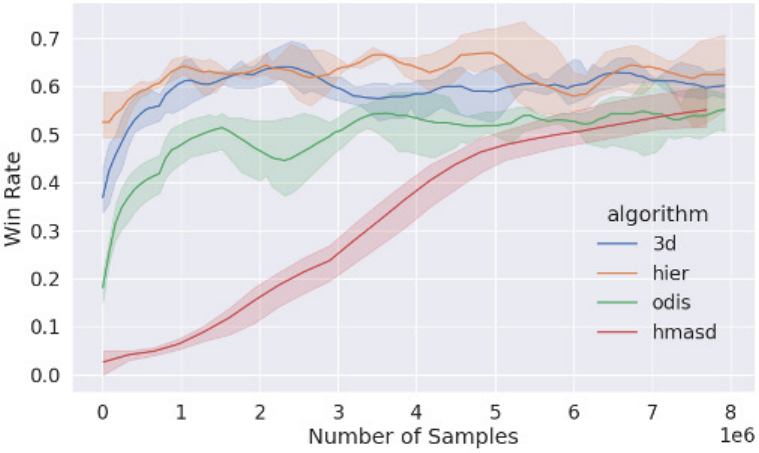}
    \label{fig:8(b)}}
    \subfigure[Terran-7]{
    \includegraphics[width=2.0in, height=1.15in]{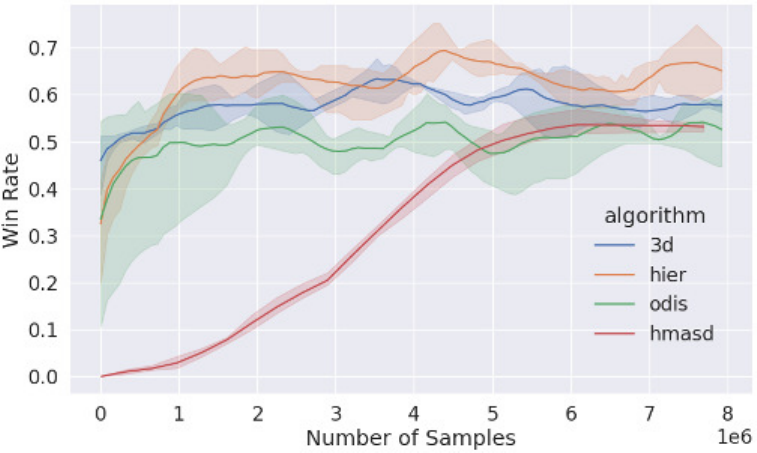}
    \label{fig:8(c)}}
    \vspace{-.1in}
\caption{Evaluation results on SMACv2. We compare the performance of online MARL using skills discovered with our methods and ODIS. As an additional baseline, we also include HMASD, an online hierarchical MARL method that discovers task-specific skills during training, as opposed to leveraging prelearned skills from offline data, as done in our methods and ODIS.}
\label{fig:8} 
\vspace{-.1in}
\end{figure*}

\begin{figure*}[t]
\centering
    \subfigure[7m]{
    \includegraphics[width=2.0in, height=1.15in]{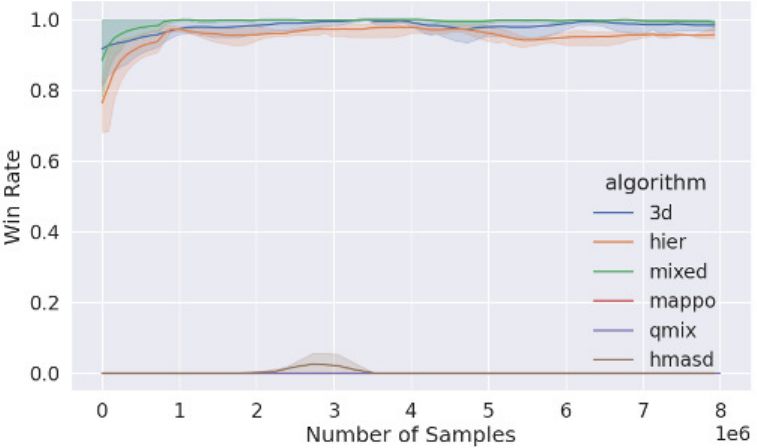}
    \label{2:fig:6(a)}}
    \subfigure[10m]{
    \includegraphics[width=2.0in, height=1.15in]{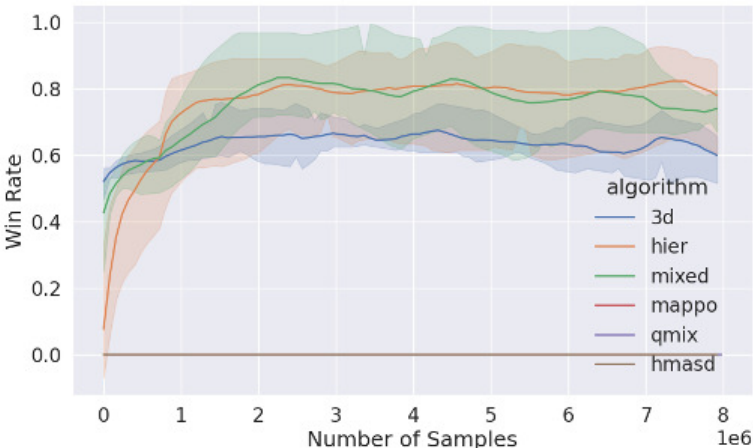}
    \label{2:fig:6(b)}}
    \subfigure[MMM2]{
    \includegraphics[width=2.0in, height=1.15in]{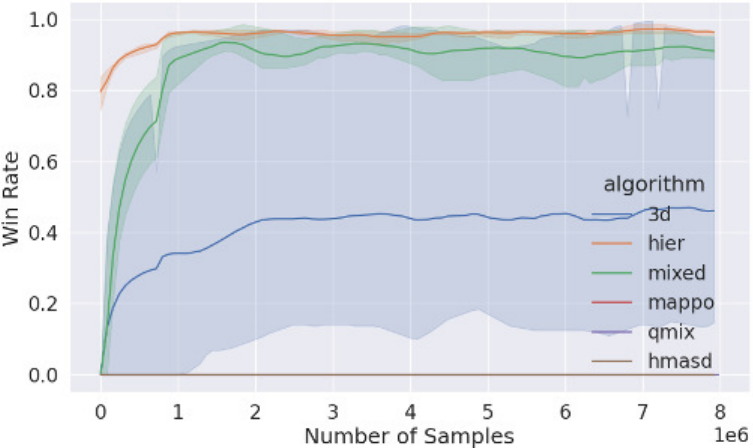}
    \label{2:fig:6(c)}}
    \vspace{-.1in}
\caption{The effectiveness of discovered skills in online MARL with sparse reward signals.}
\label{2:fig:6} 
\vspace{-.1in}
\end{figure*}

SMACv2 \cite{ellis2024smacv2} uses procedural content generation \cite{DBLP:journals/natmi/RisiT20} to address SMAC’s lack of stochasticity. In SMACv2, for each episode, team compositions and agent start positions would be generated randomly. Thus, it is no longer sufficient for agents to repeat a fixed action sequence, but they must learn to coordinate across a diverse range of scenarios. Specifically, we select the Terran task, which involves teams composed of three types of units: marines, marauders, and medivacs. As in SMAC tasks, we train an MAPPO policy as the offline data collector. However, the learned policy can only achieve a win rate around 55\% on the source tasks upon convergence, highlighting the difficulty of SMACv2. For different tasks in this task set, we vary the team size, selecting the less challenging Terran-3 and Terran-5 as source tasks and Terran-7 as the target task.

Our methods consistently outperform ODIS, discovering more effective skills for downstream MARL. 
We also adopt HMASD \cite{DBLP:conf/nips/Yang0LZL23} as a baseline. HMASD is a SOTA \textbf{online} hierarchical MARL algorithm that discovers skills while forming a hierarchical policy for a specific task. Notably, for Terran-7, the skills used by our algorithms are discovered from Terran-3 and Terran-5 and remain fixed during hierarchical policy learning, whereas HMASD develops specific skills for Terran-7. Despite this, our algorithm still achieves superior performance. As discussed in Appendix \ref{rw}, HMASD discovers only single-agent and team skills, rather than multi-agent skills for subgroups of varying sizes. Team skills can be less flexible to use, particularly when the team composition randomly changes across episodes (as in SMACv2).

\subsection{Performance in MARL with Sparse Rewards} \label{srwd}

With pretrained skills, only a high-level policy $\pi_\omega$ for skill selection needs to be trained for downstream task learning, as detailed in Alg \ref{alg:1}, and the decision horizon of $\pi_\omega$ is reduced to the original one divided by the skill length $H$. Thus, learning with skills (i.e., hierarchical learning) is particularly advantageous for long-horizon tasks with sparse and delayed reward signals. \textbf{To testify this, we modify the reward setups of the unseen tasks: 7m, 10m, MMM2, to be sparse, where agents receive a reward of 20 only upon eliminating all enemies; otherwise, they receive a reward 0.} These three tasks, with maximum episode horizons of 110, 120, and 180 respectively, are particularly challenging. 

We apply two online MARL algorithms: MAPPO \cite{yu2022surprising} and QMIX \cite{DBLP:conf/icml/RashidSWFFW18}, to these tasks, and they consistently fail with all-zero win rates. Note that we use the original code and hyperparameter setup provided in \cite{yu2022surprising}. Although they have been proposed for years, MAPPO and QMIX remain the most robust algorithms in online MARL, as verified by extensive empirical studies \cite{yu2022surprising,hu2023rethinking}. In contrast, with skills discovered using our algorithms: VO-MASD-3D, VO-MASD-Mixed, VO-MASD-Hier, the performance can be greatly improved, as shown in Figure \ref{2:fig:6}. Note that (1) skills are discovered from source tasks (rather than 7m, 10m, or MMM2) and (2) only sparse rewards are adopted for downstream online MARL. This highlights the effectiveness of hierarchical MARL when employing the multi-agent, multi-task skills discovered by our algorithms. As in Figure \ref{2:fig:3}, VO-MASD-Hier achieves the best overall performance, followed by VO-MASD-Mixed. 
Further, we compare our methods with HMASD, which discovers skills through interaction with the environment and has proven effective in sparse reward settings. However, HMASD fails in all three tasks, highlighting the superiority of the skills learned with our methods, even though they are discovered from offline data of different tasks. 
\textbf{Note that all the reported results are reproducible using the submitted code folder.}

\section{Conclusion}

In this work, we propose novel algorithms for discovering coordination patterns among agents as multi-agent skills from offline multi-task data. The key challenge lies in abstracting agents' behaviors at both the temporal and agent levels in a fully automatic manner. We address this challenge by developing novel encoder-decoder architectures and co-training the encoder-decoder with a grouping function that dynamically groups agents. Empirical results demonstrate that multi-agent skills discovered using our methods significantly enhance learning in downstream MARL tasks, particularly in scenarios with sparse reward signals. A detailed discussion of the limitations is provided in Appendix \ref{disc}.


\section*{Ethical Statement}

There are no ethical issues.

\bibliographystyle{named}
\bibliography{ijcai25}

\clearpage
\appendix
\section{Related Works} \label{rw}

In this section, we provide a comprehensive review on the use of skills in cooperative MARL, and emphasize the novelty of our proposed algorithm. We categorize these works into several groups based on their algorithm designs. The first group of works either rely on predefined skills \cite{DBLP:conf/atal/AmatoKK14,DBLP:journals/jair/AmatoKKH19} or require well-designed task hierarchies/decomposition \cite{DBLP:conf/imsccs/ShenGL06,DBLP:journals/aamas/GhavamzadehMM06,fosong2023learning}. However, we focus on automatic task decomposition and skill discovery, which is more demanding but makes our algorithm more applicable. Next, we introduce research works in this direction.


\textbf{MARL with single-agent skills:} A straightforward manner is to learn a set of single-agent skills for each agent using discovery methods proposed in single-agent scenarios \cite{DBLP:conf/iclr/EysenbachGIL19,DBLP:conf/iclr/JinnaiPMK20,DBLP:conf/nips/ChenAL23,DBLP:journals/corr/abs-2305-17327}, and then learn a multi-agent meta policy over these individual skills. The intuition is that players in a team sport can master their skills individually outside of team practices.
Specifically, the meta policy $\pi_h(\vec{z}\mid s)$ assigns skills $\vec{z}=(z^1, \cdots, z^n)$ to all agents, and then each agent decides on its primitive action according to its skill policy $\pi_l(a^i\mid \tau^i, z^i)$. Note that (1) $z^i \in \Omega^z$ is an embedding of a single-agent skill and $\Omega^z$ is a finite set of skill choices; (2) $\pi_h(\vec{z}\mid s)$ is usually implemented as $(\pi_h^1(z^1\mid \tau^1), \cdots, \pi_h^n(z^n\mid \tau^n))$ in CTDE schemes to enable decentralized skill selection. 
Representative works of this category include \cite{DBLP:conf/iclr/LeeYL20,DBLP:conf/atal/YangBZ20,DBLP:conf/gecco/SachdevaKMT21}. In \cite{DBLP:conf/iclr/LeeYL20}, skills $\pi_l$ are learned in a separate stage, while, in \cite{DBLP:conf/atal/YangBZ20,DBLP:conf/gecco/SachdevaKMT21}, skills are concurrently trained with the meta policy $\pi_h$. Typically, as in single-agent scenarios, the skill duration $H$ is a predefined value, and a new skill assignment for all agents, i.e., $\vec{z}$, should be given by the meta policy every $H$ time steps. However, some algorithms \cite{DBLP:conf/atal/ChakravortyWRCB20,DBLP:conf/nips/XiaoTA22,DBLP:conf/kdd/Zhang0LW023} have been proposed for the case where the skills of each agent can take different amounts of time and so the skill selection across agents can be asynchronized. 
To sum up, this group of works replace the primitive action set $A$ in MARL with an individual-skill set $\Omega^z$, which could simplify the learning especially for long-horizon tasks. 
However, in multi-agent scenarios, discovering inter-agent coordination patterns as multi-agent skills is possible. Learning with multi-agent skills could be simpler, since they constitute higher-level abstractions of multi-agent behaviors than single-agent skills and are closer in form with the overall multi-agent policy.

\textbf{Role-based MARL:} Another main branch of algorithms is role-based MARL. These algorithms, based on the CTDE scheme, usually contain three modules: establishing role representations $\Omega^Z = \{Z_1, \cdots, Z_m\}$, learning a role selector $\pi_h(Z^i\mid \tau^i)$ 
($Z^i \in \Omega^Z$), and learning role policies $\pi_l(a^i\mid \tau^i, Z^i)$.
This framework is similar with the one used for MARL with single-agent skills. However, the policy of role $Z$, i.e., $\pi_l(a^i\mid \tau^i, Z)$, is not a single-agent skill policy but a policy for the subgroup $g_Z$. This is because each agent $i \in g_Z$ adopts the same role policy $\pi_l(\cdot\mid \cdot, Z)$ and $\pi_l(\cdot\mid \cdot, Z)$ is trained in a centralized manner with the aim for the subgroup $g_Z$ to maximize a global return. As a comparison, aforementioned \cite{DBLP:conf/iclr/LeeYL20,DBLP:conf/gecco/SachdevaKMT21} learn skill policies $\pi_l(\cdot\mid \cdot, z)$ based on reward functions specifically defined for single-agent skills, and \cite{DBLP:conf/atal/YangBZ20} updates skill policies through Independent Q-learning, i.e., a fully-decentralized training scheme, thus the learned skills are for individuals. Notable works in this category, roughly ordered by the publication date, include \cite{DBLP:conf/icml/0001DLZ20,DBLP:conf/icml/LiuLSGZA21,DBLP:conf/iclr/00010MPWZ21,DBLP:conf/nips/LiuLXDL22,DBLP:conf/nips/IqbalCS22,DBLP:conf/nips/Yang0HZZL22,li2023celebrating,DBLP:conf/nips/ZangHLFFXC23,DBLP:conf/nips/TianCHLZWPGDGC23,DBLP:conf/aaai/XuBZ0F23,xia2023dynamic,zhou2024constructing}. Among these works, $Z$ is given different names, such as role \cite{DBLP:conf/icml/0001DLZ20}, ability \cite{DBLP:conf/nips/Yang0HZZL22}, subtask \cite{li2023celebrating}, and skill \cite{DBLP:conf/nips/LiuLXDL22}, but refers to the same concept. All these works utilize a similar algorithm framework, which, as mentioned above, contains three modules for learning the role embedding, role selector\footnote{Most algorithms in this category adopt decentralized selectors, i.e., $(\pi_h(Z^1\mid \tau^1), \cdots, \pi_h(Z^n\mid \tau^n))$, but there are some works, such as \cite{DBLP:conf/icml/LiuLSGZA21,DBLP:conf/nips/IqbalCS22}, utilizing centralized ones, i.e., $\pi_h(\vec{Z}\mid s)$. 
The global state could provide more information for the coordinated role assignment.}, and role policy, respectively. One main distinction lies in their varied approaches for learning the role embedding, which can be based on action effects \cite{DBLP:conf/iclr/00010MPWZ21}, global state reconstruction \cite{zhou2024constructing}, or predictions of the next observation and reward \cite{DBLP:conf/nips/LiuLXDL22}. 
As a recommendation, readers who are new to this area can refer to the two representative works: \cite{DBLP:conf/nips/Yang0HZZL22,DBLP:conf/aaai/XuBZ0F23}. 
To sum up, although the role policy $\pi_l(a^i\mid \tau^i, Z)$ is more than a single-agent skill and learned as a subgroup policy for $g_Z$, agents in $g_Z$ are similar in behaviors as they all adopt a policy conditioned on $Z$. As mentioned in 
\cite{DBLP:conf/nips/Yang0HZZL22}, role-based MARL is designed to group agents with similar abilities into the same subtask. However, multi-agent skills should be abstractions of subgroup coordination patterns, and agents from this subgroup could possess heterogeneous behaviors. For instance, the collaboration between two pilots -- one proficient in advanced flying maneuvers and the other in weapon control -- while operating a fighter jet, exemplifies a multi-agent skill. Therefore, the concept of a multi-agent skill is more generalized than that of role policy and cannot be acquired through aforementioned role-based algorithms.

\textbf{Team skill discovery:} There are relatively few works on multi-agent skill discovery. The authors of \cite{DBLP:journals/corr/abs-2006-04021,DBLP:conf/nips/ChenCLA22,DBLP:conf/nips/Yang0LZL23} propose algorithms to discover skills for the entire team of agents. As a representative, in \cite{DBLP:conf/nips/Yang0LZL23}, they adopt a transformer-based skill selector $\pi_h(z^{1:n}, Z\mid s)$ to decide on the team skill $Z$ and individual skills $z^{1:n}$ autoregressively
based on the global state. Then, each agent $i$ interacts with the environment using a corresponding policy $\pi_l(a^i\mid \tau^i, z^i, Z)$\footnote{In \cite{DBLP:conf/nips/Yang0LZL23}, $Z$ is not used as a condition of $\pi_l$ but adopted for centralized training, so the team skill information is implied in $\pi_l$.}. Compared with the role policy $\pi_l(\cdot\mid \cdot, Z)$, the team skill $\pi_l(\cdot\mid \cdot, z, Z)$ could contain heterogeneous behaviors across agents, which are embedded as various individual skills $z^{1:n}$. However, the team skill is only a special instance of multi-agent skills, as the number of agents within a team skill is always $n$. 
Ideally, multi-agent skill discovery should identify subgroups where agents interact frequently and extract their behavior patterns as joint skills, and the size of the subgroup could vary from 1 to $n$, which is much more challenging as it additionally requires dynamic grouping according to the task scenario. 

All the algorithms mentioned above are for \textbf{online} skill discovery, while the authors of \cite{DBLP:conf/iclr/ZhangJLY0Z23} propose an approach for discovering coordination skills from offline data. However, this algorithm is still a role-based one, 
 and the learned role policy $\pi_l(\cdot\mid \cdot, Z)$ is different from multi-agent skills in concept as mentioned above. The difference between \cite{DBLP:conf/iclr/ZhangJLY0Z23} and aforementioned role-based methods is that it replaces task rewards with the reconstruction accuracy of joint actions, so that the learned skills are not task-specific but generalizable. 

In conclusion, research on multi-agent skill discovery is still at an early stage of development, especially for the offline setting. Even without prelearned skills, when dealing with a complex multi-agent task, the agents would implicitly learn to decompose the overall task into several subtasks, assign a subgroup for each subtask, and develop a joint policy (i.e., multi-agent skill) within the subgroup to handle the corresponding subtask. Replacing primitive actions with single-agent skills or role policies could make such a learning process more efficient, as agents can assemble these higher-level abstractions to obtain the required subgroup joint policies more easily. As the first offline multi-agent skill discovery algorithm, our work takes one step further by directly identifying subgroups, which could change throughout a decision horizon, and extracting their coordination patterns as multi-agent skills. With these joint skills, the MARL process could be greatly simplified, since agents only need to select correct multi-agent skills without considering grouping with others or forming subgroup policies. Compared to single-agent skills or role policies, multi-agent skills make better use of the offline multi-agent interaction data, representing a more efficient form of knowledge discovery. 


\section{Alternative Approaches for Matching Skill Embeddings with the Codebook} \label{GAMSEC}

The approach for mapping \( z^{1:n} \) to \( e^{1:n} \), as shown in Figure \ref{fig:ctde}, utilizes the prelearned grouper $h_\psi$, which requires global state $s$ and previous agents' grouping result $g^{1:i-1}$ to decide on $g^i$. Global information can facilitate subgroup division, but it may not be accessible during execution, for which we have several solutions.\footnote{
Previous works on task decomposition in MARL \cite{DBLP:conf/iclr/ZhangJLY0Z23,DBLP:conf/nips/Yang0LZL23} also rely on global information; therefore, our algorithm design does not introduce any additional requirements.} First, during the offline skill discovery stage, we could replace $s$ with the local observation $o^i$ (or observation history) as the input of $h_\psi$, shown in Figure \ref{fig:1} (a). In this way, $s$ is not required during execution. This replacement would lead to information loss for grouping. However, we note that $h_\psi$ is trained with MAPPO, which involves a centralized critic $V_{\eta'}(s)$ to guide the learning with global information, so the information required for grouping is implicitly involved. Second, we propose a greedy algorithm to directly match $z^{1:n}$ with the codebook, which does not use $h_\psi$ or additional input other than $z^{1:n}$.

\begin{algorithm}[t] 
\caption{Multi-agent skill assignment} \label{2:alg:2}
\begin{algorithmic} 
    \STATE Input: $z^{1:n}$, $E_{1:n}$
    \STATE Initialize a Min-Heap $M$
    \FOR{$i=1\cdots n$}
    \FOR{each $i$-agent subgroup $\vec{j}$}
    \FOR{each $i$-agent code $\vec{e}$ in $E_i$}
    \STATE Insert $(\|z^{\vec{j}} - \vec{e}\|_2^2/i, \vec{j}, \vec{e})$ into $M$
    \ENDFOR
    \ENDFOR
    \ENDFOR
    \WHILE{$i < n$ ($i$ is initialized as 0)}
    \STATE $d, \vec{j}, \vec{e} \leftarrow M.\text{pop()}$
    \IF{all agents in $\vec{j}$ remain unassigned}
    \STATE $e^{\vec{j}} \leftarrow \vec{e}$, $i \mathrel{+}= |\vec{j}|$
    \ENDIF
    \ENDWHILE
    \STATE Return $e^{1:n}$
\end{algorithmic}
\end{algorithm}

The rule-based multi-agent skill assignment process is shown as Alg \ref{2:alg:2}. Given skill embeddings $z^{1:n}$ produced by the high-level policy $\pi_\omega$, we repeat the following process until all agents are assigned with skills: greedily select the closest multi-agent code $\vec{e}$, assign the corresponding multi-agent skill to the selected subgroup $\vec{j}$, remove this subgroup from the waiting list. We can implement such process with a Min-Heap, from which we can efficiently query the closet pair of skill embeddings and codes (via the ``pop" operation).

\begin{table*}[t]
\centering
\caption{Descriptions of the marine and MMMs task sets}
\begin{tabular}{|c|c|c|c|c|c|}
\hline
{Task Set} & {Property} & {Task} & {Type} & {Ally Units} & {Enemy Units} \\
\hline
\hline
\multirow{4}{*}{marine} & \multirow{4}{*}{\makecell{homogeneous, \\ symmetric}} & 3m & source & 3 marines & 3 marines\\ 
                      &  & 5m & source & 5 marines & 5 marines \\ \cline{3-6}
                      &  & 7m & unseen & 7 marines & 7 marines \\
                      &  & 10m & unseen & 10 marines & 10 marines \\ \hline
\multirow{2}{*}{MMMs} & \multirow{2}{*}{\makecell{heterogeneous, \\ asymmetric}} & MMM & source & \makecell{1 mv, 2 md, 7 mn} & \makecell{1 mv, 2 md, 7 mn}\\ \cline{3-6}
                      &  & MMM2 & unseen & \makecell{1 mv, 2 md, 7 mn} & \makecell{1 mv, 3 md, 8 mn} \\
\hline
\end{tabular}
\label{2:table:1}
\end{table*}

\begin{figure*}[t]
\centering
\includegraphics[width=5.5in, height=1.5in]{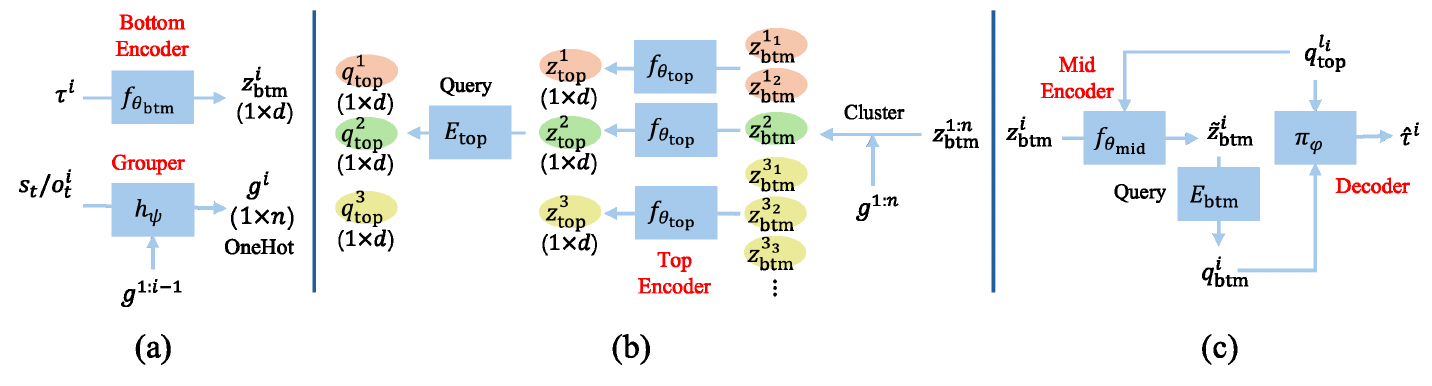}
\caption{An alternative design of VO-MASD-Hier.}
\label{2:fig:5} 
\end{figure*}

To make full use of the discovered joint skills, instead of independently selecting a skill code (i.e., a $1 \times d$ single-agent code from a complete $m \times d$ code) for each agent, we can assign each multi-agent code as a whole to preserve the coordination pattern embedded within the multi-agent code, which motivates the design of Alg \ref{2:alg:2}. As a remark, we provide comparisons with related works: (1) Our $(m \times d)$ multi-agent codes embed coordination patterns among $(m)$ agents. Each agent's behavior is embedded by a single-agent $(1 \times d)$ code, allowing for individual differences. In contrast, role-based algorithms learn a shared role policy for subgroups of agents with similar behaviors and abilities. (2) Alg \ref{2:alg:2} only requires fairly compact centralized information, i.e., $z^{1:n}$, for coordinated skill assignment, where each skill embedding $z^i$ is decided based on local observations of agent $i$ rather than global information (e.g., $s$) as in \cite{DBLP:conf/icml/LiuLSGZA21,DBLP:conf/nips/Yang0LZL23}. It's also worth noting that VO-MASD-Mixed (introduced in Section \ref{VO-MASD-3D}) does not require any global information (e.g., $s$ or $z^{1:n}$) for the skill assignment, since each agent selects its skill independently.

However, as a limitation, when $n$ is large, Alg \ref{2:alg:2} can be inefficient. For solutions, we can (1) avoid discovery of $x$-agent skills, where $x$ is around $n/2$, as the combination number $\binom{n}{k}$ could be large; (2) utilize domain knowledge to filter out useless skill codes in $E_{1:n}$ or specify the size of subgroups, i.e., $x$. 
Further, we note that Alg \ref{2:alg:2} is a greedy assignment method which would inevitably bring suboptimality.


\section{Details of the SMAC Task Sets} \label{SMACT}

The marine task set includes four marine battle tasks, for each of which several ally marines need to beat the same number of enemy marines to win; while in the MMMs task set, each task is a battle between two groups of medivacs (mv), marauders (md), and marines (mn). Detailed descriptions of these task sets are listed in Table \ref{2:table:1}. We note that skills are discovered from source tasks in a task set and evaluated on both source and unseen tasks. Given that MMM2 is categorized as super-hard in SMAC \cite{DBLP:conf/atal/SamvelyanRWFNRH19}, skills discovered solely in MMM would fail in MMM2, regardless of the skill discovery method employed. For  effective comparisons, we instead use a mixture of offline data from both MMM and MMM2 to discover skills, with MMM2 trajectories constituting less than 5\% of the total.

\section{An Alternative Design of VO-MASD-Hier} \label{ADVMH}

The only difference between Figures \ref{fig:2} and \ref{2:fig:5} is in part (c). An extra encoder $f_{\theta_{\text{mid}}}$ is introduced to further embed $z^i_{\text{btm}}$ and its corresponding top code $q^{l_i}_{\text{top}}$ to a bottom skill embedding $\Tilde{z}^i_{\text{btm}}$, which is then used to query a bottom code $q^i_{\text{btm}}$, while in VO-MASD-Hier, $z^i_{\text{btm}}$ is directly matched with $E_{\text{btm}}$ for $q^i_{\text{btm}}$. This three-level encoder design is used in VQ-VAE-2 \cite{DBLP:conf/nips/RazaviOV19} which demonstrates superior performance for image generation. However, this alternative design underperforms VO-MASD-Hier in multi-agent skill discovery, as shown in Figure \ref{2:fig:4(a)}. One possible explanation is that the inductive bias: primitive actions $\rightarrow$ single-agent skills $\rightarrow$ multi-agent skills, is not well-adopted in this design. Specifically, in Figure \ref{2:fig:5} (c), the single-agent skill embedding $\Tilde{z}^i_{\text{btm}}$ involves information from the multi-agent skill code $q^{l_i}_{\text{top}}$, which violates the aggregation direction: single-agent skills $\rightarrow$ multi-agent skills. In contrast, VO-MASD-Hier eliminates $f_{\theta_{\text{mid}}}$ for a simpler architecture while adhering to this bias.

\section{Compute Resources}

Experiments were conducted using the Oracle Cloud infrastructure, where each computation instance was equipped with an NVIDIA Tesla P100 GPU, 12 Intel Xeon Platinum CPU cores, and 72 GB of memory. For VO-MASD-Mixed, VO-MASD-3D, and VO-MASD-Hier, each instance could simultaneously handle one of these experiment sets: [terran-3, terran-5, terran-7], [3m, 5m, 7m, 10m], [MMM], or [MMM2]. While, for ODIS, VO-MASD-Single, MAPPO, QMIX, and HMASD, the capacity of each instance could be doubled. The average running time of each experiment set is approximately 2 days. For example, without conducting repeated runs (utilizing different random seeds), the experiments depicted in Figure \ref{2:fig:3}, Figure \ref{2:fig:4}, and Figure \ref{2:fig:5} require approximately 600, 400, and 250 GPU hours, respectively, using such computation instance. The full research project required more compute than the experiments reported in the paper, primarily for the hyperparameter tuning and testing of alternative designs. Different hyperparameter configurations could lead to better or worse performance, but we suggest using the provided ones in the released code folder.

\section{Additional Evaluation Results}

Due to the page limit, we present additional empirical results in this section as a supplement to the main findings discussed in Section \ref{eval}.

\begin{figure*}[t]
\centering
    \subfigure[odis-10m-1]{
    \includegraphics[width=1.6in, height=1.55in]{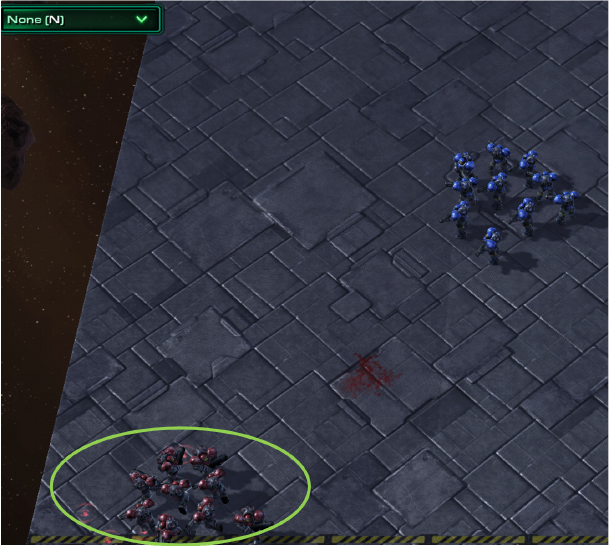}
    \label{fig:6(e)}}
    \subfigure[odis-10m-2]{
    \includegraphics[width=1.6in, height=1.55in]{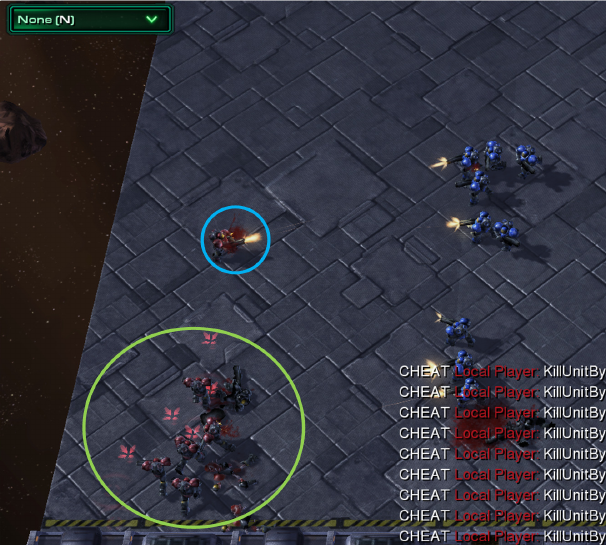}
    \label{fig:6(f)}}
    \subfigure[odis-MMM2-1]{
    \includegraphics[width=1.6in, height=1.55in]{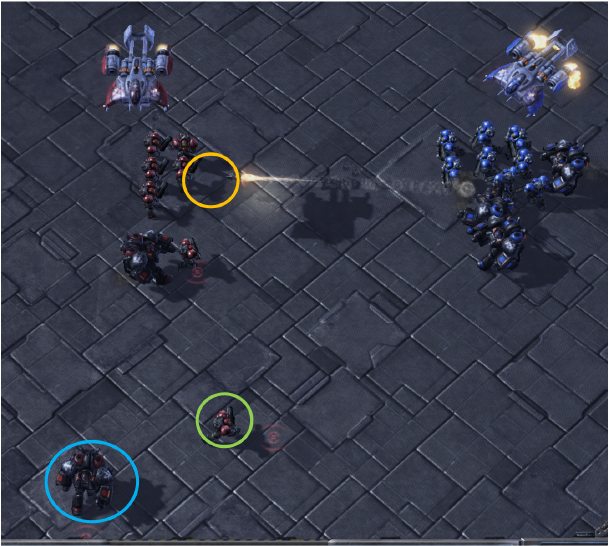}
    \label{fig:6(g)}}
    \subfigure[odis-MMM2-2]{
    \includegraphics[width=1.6in, height=1.55in]{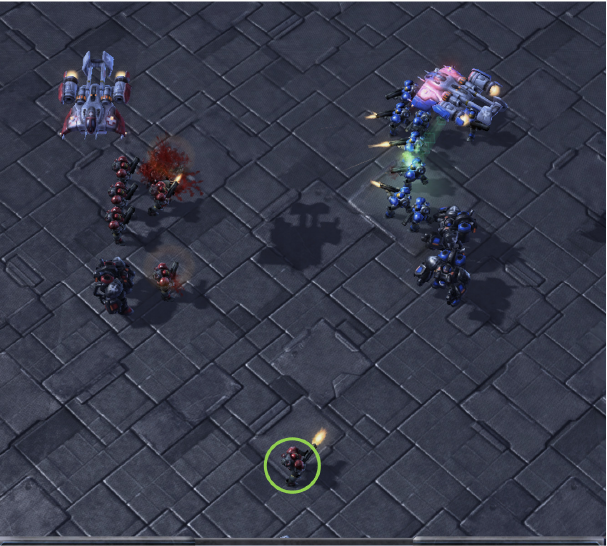}
    \label{fig:6(h)}}
    \subfigure[3d-10m-1]{
    \includegraphics[width=1.6in, height=1.55in]{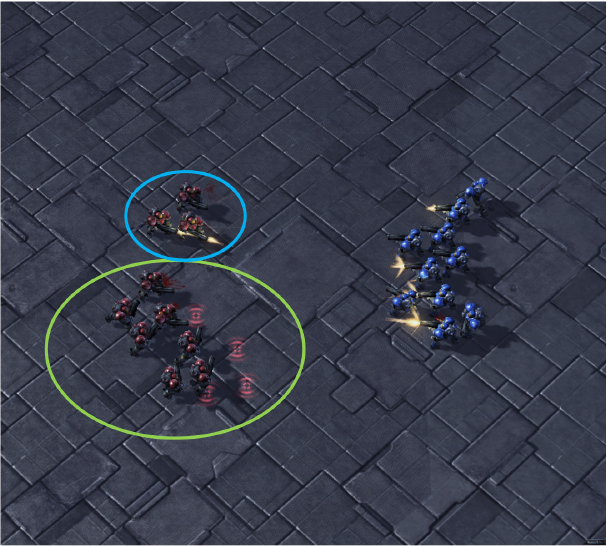}
    \label{fig:6(a)}}
    \subfigure[3d-10m-2]{
    \includegraphics[width=1.6in, height=1.55in]{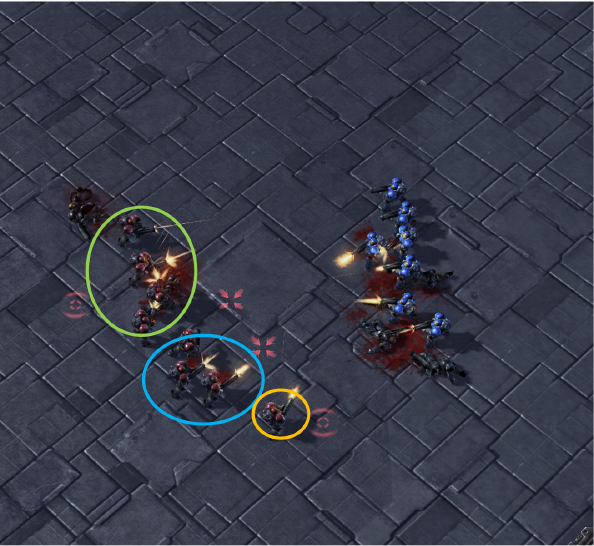}
    \label{fig:6(b)}}
    \subfigure[3d-MMM2-1]{
    \includegraphics[width=1.6in, height=1.55in]{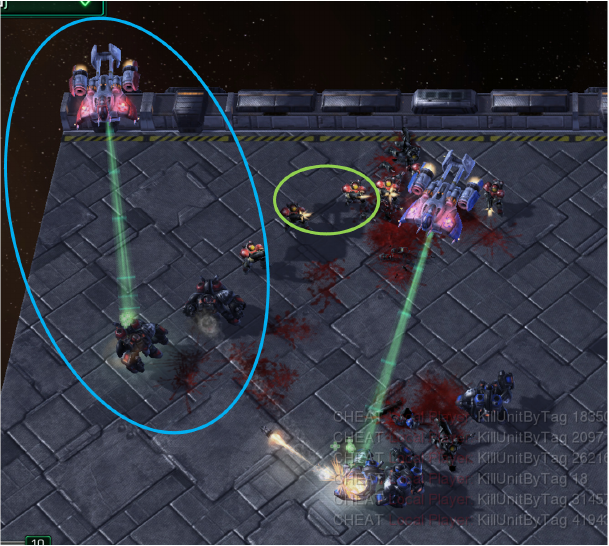}
    \label{fig:6(c)}}
    \subfigure[3d-MMM2-2]{
    \includegraphics[width=1.6in, height=1.55in]{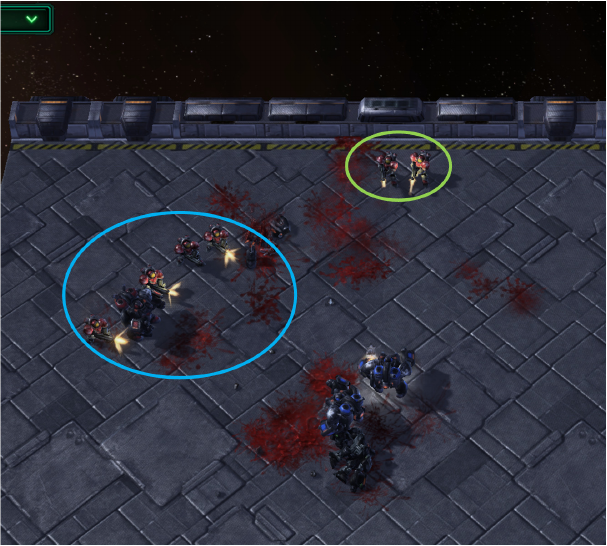}
    \label{fig:6(d)}}
    \subfigure[hier-10m-1]{
    \includegraphics[width=1.6in, height=1.55in]{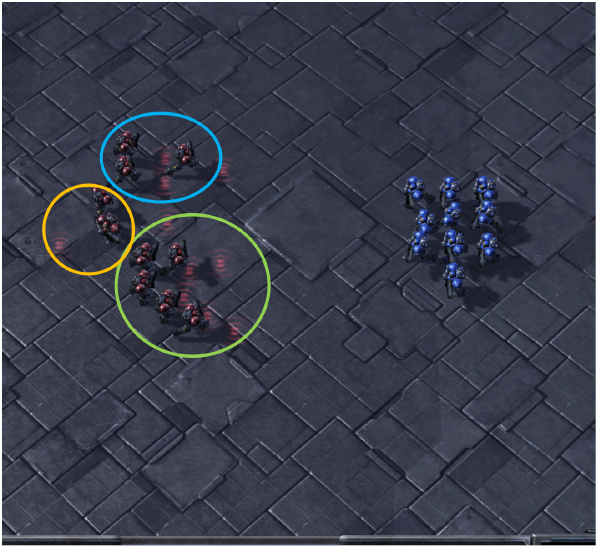}
    \label{fig:6(i)}}
    \subfigure[hier-10m-2]{
    \includegraphics[width=1.6in, height=1.55in]{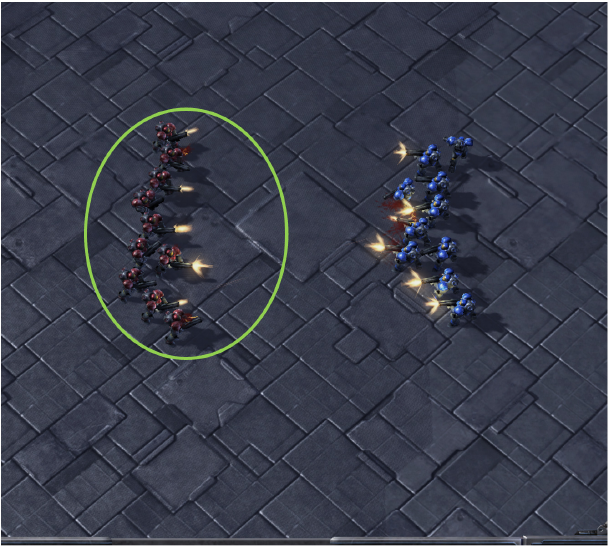}
    \label{fig:6(j)}}
    \subfigure[hier-MMM2-1]{
    \includegraphics[width=1.6in, height=1.55in]{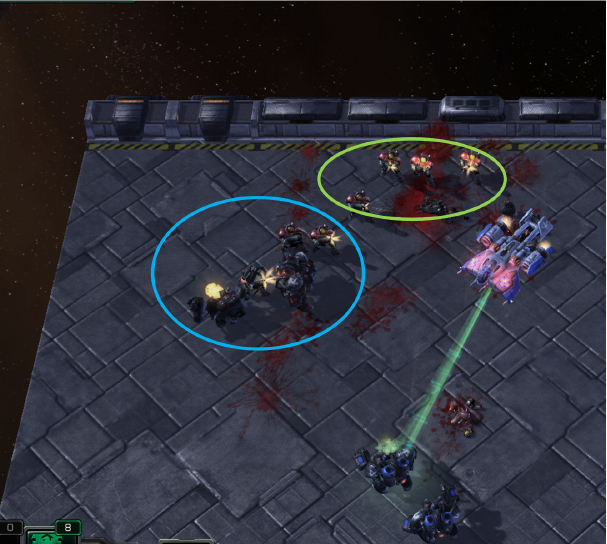}
    \label{fig:6(k)}}
    \subfigure[hier-MMM2-2]{
    \includegraphics[width=1.6in, height=1.55in]{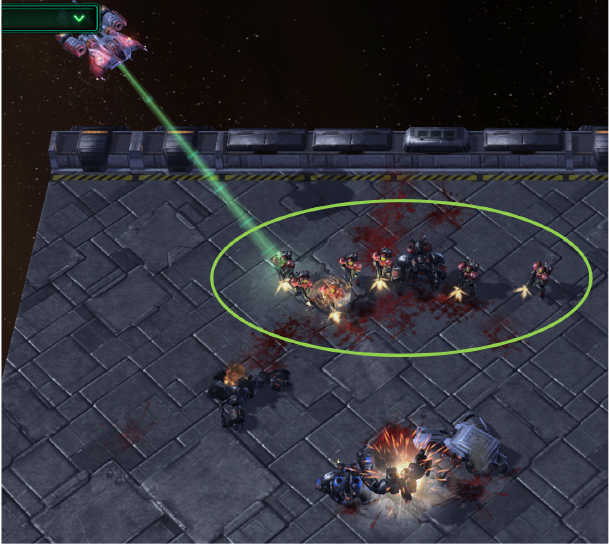}
    \label{fig:6(l)}}
\caption{Visualization of the skills learned using ODIS, VO-MASD-3D, and VO-MASD-Hier on the two most challenging SMAC tasks: 10m and MMM2. The circles in the figure represent subgroups of agents.}
\label{fig:6} 
\end{figure*}

\subsection{Qualitative Analysis of Discovered Skills} \label{QADS}

In Figure \ref{fig:6}, we show representative behaviors of the policies learned using ODIS, VO-MASD-3D, and VO-MASD-Hier in the challenging, unseen tasks: 10m and MMM2. A qualitative analysis of these results can provide insights into whether and how effectively the algorithms capture coordination patterns among agents (from source task demonstrations) and leverage them to develop effective team strategies.

For ODIS, the agents exhibit homogeneous behaviors, reflecting the characteristics of role-based MARL methods. For instance, in Figures \ref{fig:6(e)} and \ref{fig:6(f)}, most agents choose to move to the corner at the beginning of an episode—a strategy likely inherited from the 3m and 5m tasks, but ineffective in the 10m task. Moreover, we do not observe effective subgroup coordination patterns: as shown in Figures \ref{fig:6(f)} through \ref{fig:6(h)}, agents attack or move independently, and the medivacs fail to heal their teammates during attacks in MMM2.

In contrast, efficient coordination among agents emerges using the skills discovered by our methods. In Figures \ref{fig:6(a)} and \ref{fig:6(b)}, one subgroup initiates an attack while another subgroup moves to form a fan-shaped formation. Subsequently, the agents split into smaller subgroups to target different enemies. 
In Figures \ref{fig:6(c)} and \ref{fig:6(d)}, the subgroup labeled with the green circle strategically attacks the opposing team's medivac first and targets an enemy on the ground after destroying the medivac. Meanwhile, the blue circle subgroup, composed of marauders, marines, and a medivac, employs distinct individual skills (as a multi-agent skill) to eliminate most of the ground enemies. We observe similar patterns in Figures \ref{fig:6(i)} - \ref{fig:6(l)}, but compared to VO-MASD-3D agents, VO-MASD-Hier agents achieve a greater numerical advantage over the opposing team, which may explain its superior performance in the super hard task MMM2.

\begin{figure*}[t]
\centering
    \subfigure[7m-med]{
    \includegraphics[width=2.2in, height=1.4in]{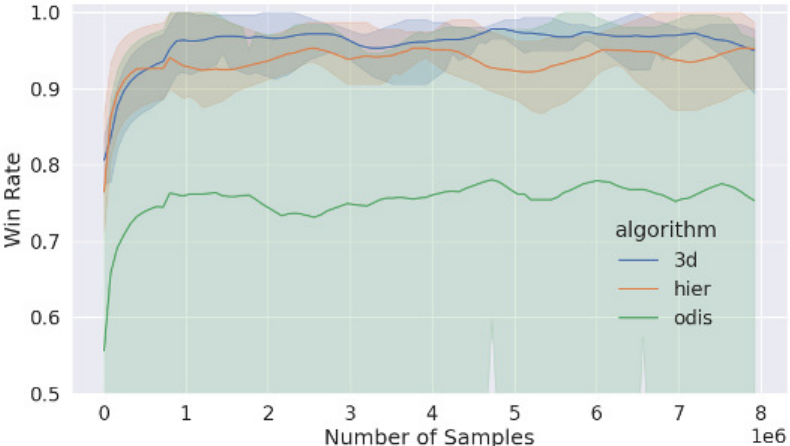}
    \label{fig:7(a)}}
    \subfigure[7m-mixed]{
    \includegraphics[width=2.2in, height=1.4in]{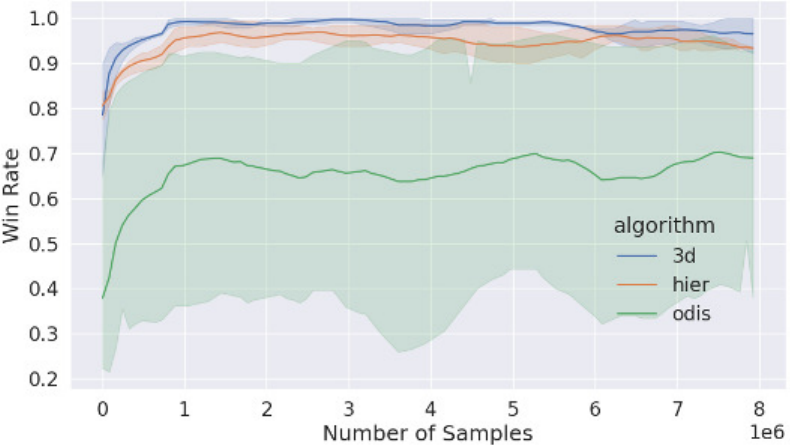}
    \label{fig:7(b)}}
    \subfigure[7m-exp]{
    \includegraphics[width=2.2in, height=1.4in]{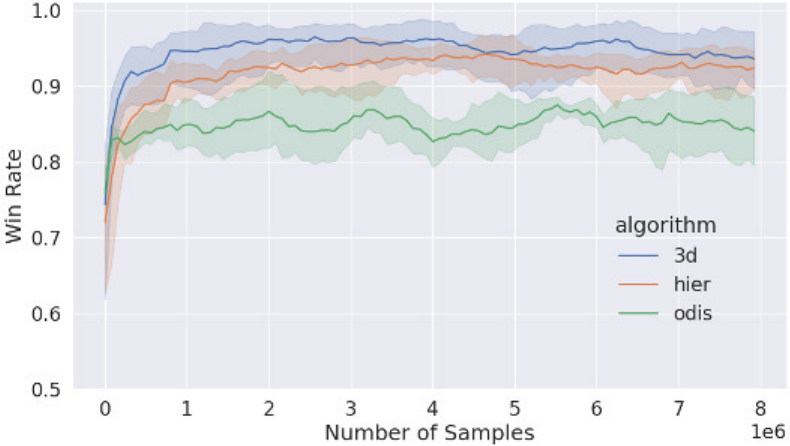}
    \label{fig:7(c)}}
    \subfigure[10m-med]{
    \includegraphics[width=2.2in, height=1.4in]{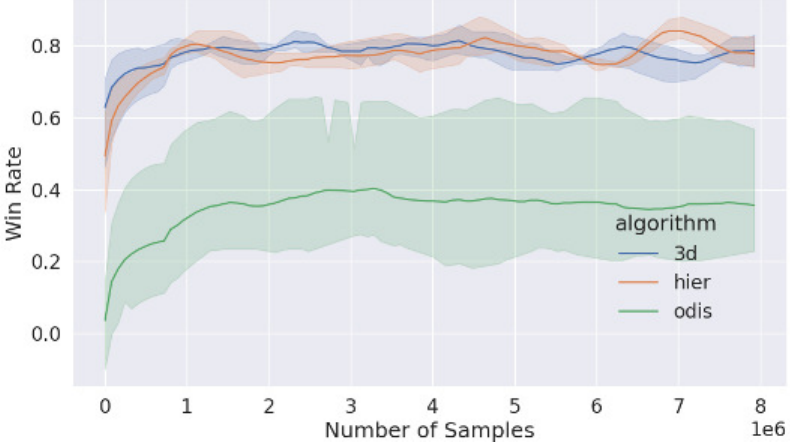}
    \label{fig:7(d)}}
    \subfigure[10m-mixed]{
    \includegraphics[width=2.2in, height=1.4in]{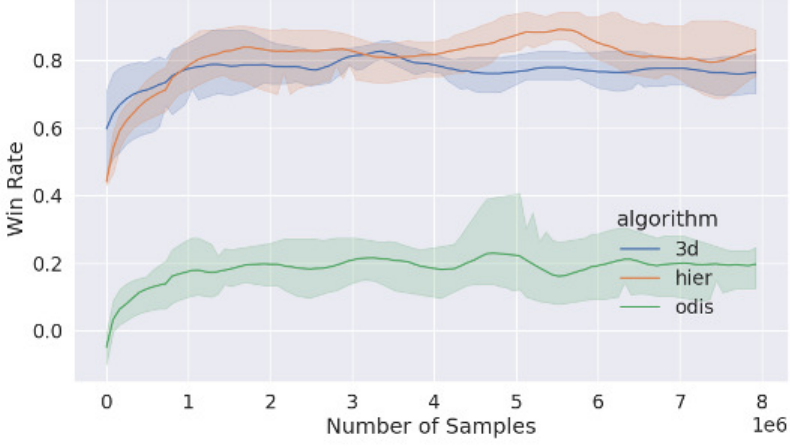}
    \label{fig:7(e)}}
    \subfigure[10-exp]{
    \includegraphics[width=2.2in, height=1.4in]{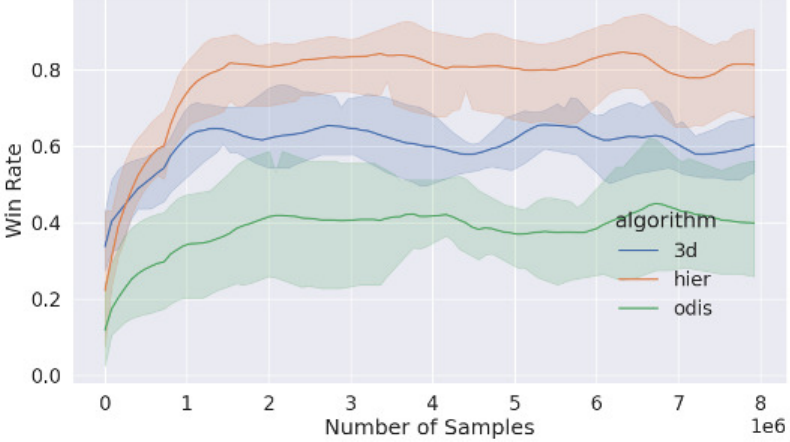}
    \label{fig:7(f)}}
    \subfigure[MMM2-med]{
    \includegraphics[width=2.2in, height=1.4in]{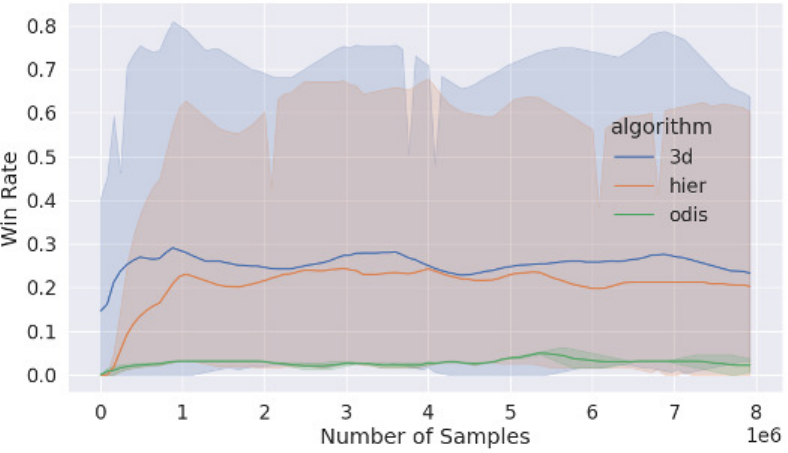}
    \label{fig:7(g)}}
    \subfigure[MMM2-mixed]{
    \includegraphics[width=2.2in, height=1.4in]{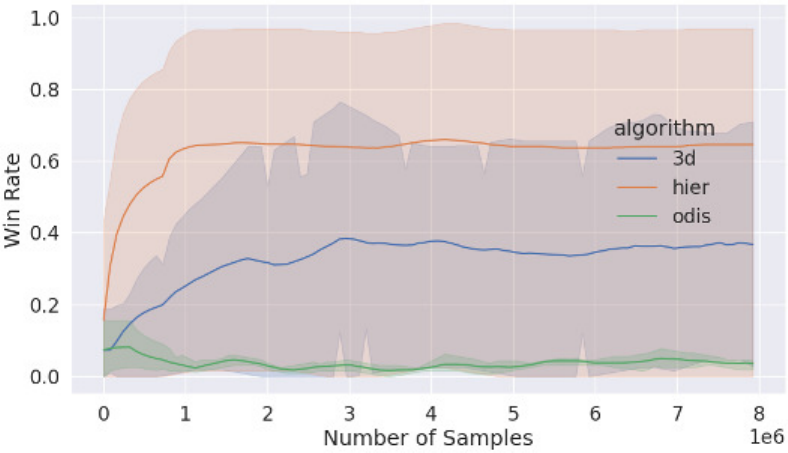}
    \label{fig:7(h)}}
    \subfigure[MMM2-exp]{
    \includegraphics[width=2.2in, height=1.4in]{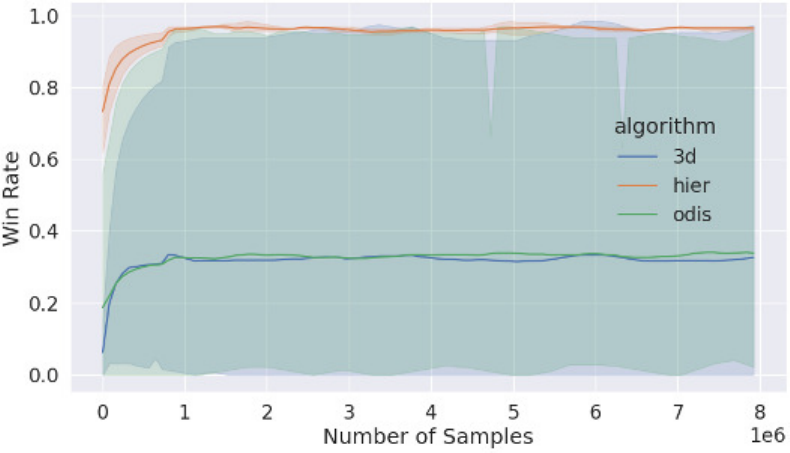}
    \label{fig:7(i)}}
\caption{Comparisons of the online MARL performance on unseen tasks using skills learned with our methods and ODIS. The skills are discovered from offline data of the source tasks, with each column corresponding to data of certain quality. Specifically, `med' represents medium-level, `exp' represents expert-level, and `mixed' is a combination (50\%-50\%) of medium and expert levels of data.}
\label{fig:7} 
\end{figure*}

\subsection{The Influence of Demonstration Quality} \label{IODQ}

As an extension of the results shown in Figure \ref{2:fig:3}, we evaluate the offline skill discovery algorithms: VO-MASD-3D, VO-MASD-Hier, and ODIS using demonstrations of varying qualities. The results are shown in Figure \ref{fig:7}. Specifically, we select an MAPPO policy with approximately a 60\% win rate in the source task (i.e., 3m, 5m, and MMM) to generate medium-level demonstrations (labeled as `med'). The offline data used in Figure \ref{2:fig:3} is considered expert-level (labeled as `exp'). A combination of these two datasets, of equal size, is labeled as `mixed'. Compared to ODIS, our methods show greater robustness to demonstration quality, particularly in 7m and 10m. Interestingly, VO-MASD-3D performs better on medium-level data than on expert-level data. This could be because, although the data are medium for the source tasks (3m and 5m), they contain more useful patterns for the target task 10m. Also, we observe significant performance variation on the super hard task MMM2 across different random seeds. However, the demonstrations were sampled using a single random seed. It appears that pattern diversity within the demonstration is crucial for robust downstream performance of the discovered skills.

\subsection{Ablation Study} \label{abl}

\begin{figure*}[t]
\centering
    \subfigure[7m]{
    \includegraphics[width=2.2in, height=1.4in]{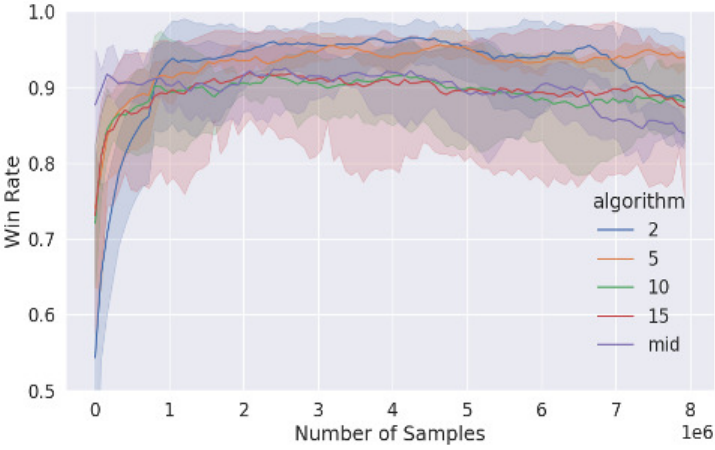}
    \label{2:fig:4(a)}}
    \subfigure[7m]{
    \includegraphics[width=2.2in, height=1.4in]{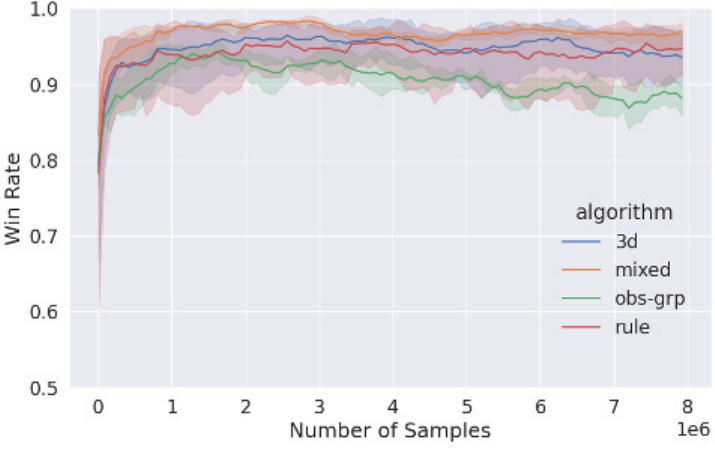}
    \label{2:fig:4(b)}}
    \subfigure[10m]{
    \includegraphics[width=2.2in, height=1.4in]{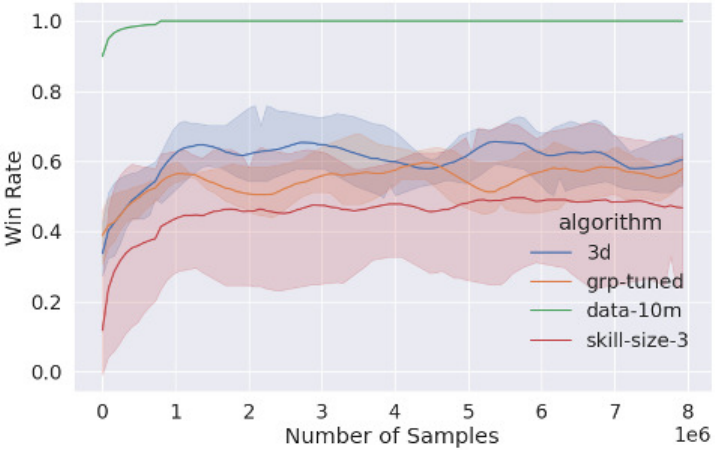}
    \label{2:fig:4(c)}}
\caption{(a) The performance of VO-MASD-Hier with different skill horizons (2--15) or an extra skill encoder $f_{\theta_{\text{mid}}}$ on 7m; (b) Comparisons among the four utilization manners of skills discovered by VO-MASD-3D on 7m; (c) Potential approaches to improve VO-MASD-3D's performance on 10m.}
\label{2:fig:4} 
\end{figure*}

Finally, we show some ablation study results in Figure \ref{2:fig:4}. In (a), we compare the performance of VO-MASD-Hier with skills of different lengths (i.e., 2, 5, 10, 15) on task 7m, where our setup (i.e., $H=5$) performs the best. Utilizing skills of length 15 causes inflexibility and inferior performance, since these skills are learned from 3m and 5m and not updated during downstream online MARL. However, using such long skills could effectively decrease the decision horizon of the high-level policy, while remaining reasonable performance which is better than the ones of ODIS and VO-MASD-Single (as shown in Figure \ref{2:fig:3(c)}). Additionally, we compare VO-MASD-Hier with its alternative design (labelled as `mid'), which adopts an extra encoder $f_{\theta_{\text{mid}}}$ to get bottom skill embeddings and is further detailed in Appendix \ref{ADVMH}. This alternative design is closer in form with VQ-VAE-2 which inspires VO-MASD-Hier. This algorithm has better initial performance but converges at a lower level. 

In (b), we compare the four utilization manners of multi-agent skills discovered by VO-MASD-3D, as detailed in the last paragraph of Section \ref{VO-MASD-3D}. `3d' and `mixed' correspond to VO-MASD-3D and VO-MASD-Mixed, respectively. `rule' refers to rule-based skill selection (i.e., Alg \ref{2:alg:2}), while `obs-grp' denotes using a grouping function $h_\psi$ that depends on $o_t^i$ instead of $s_t$ (see Figure \ref{fig:1}). Notably, `mixed', `obs-grp', and `rule' do not rely on states during execution as `3d'.  `rule' and `mixed' have comparable or even better performance compared to `3d', and the inferior performance of `obs-grp' could potentially be improved by relacing $o_t^i$ with the observation-action history. 

In (c), we explore some approaches to improve the performance of VO-MASD-3D on unseen tasks (e.g., 10m). In the original setup (i.e., `3d'), skills discovered from 3m and 5m are $m$-agent coordination patterns, where $m$ ranges from 1 to 5. `skill-size-3' limits $m$ to a range of 1 to 3, corresponding to scenarios where domain knowledge is available and we only need to learn skills for specific subgroup sizes. However, its worse performance (compared to `3d') shows the necessity to utilize skills of large subgroups for this task. The grouper $h_\psi$ as shown in Figures \ref{fig:1} and \ref{fig:2} is trained in a multi-task manner (e.g., in 3m and 5m)\footnote{The observation, state, and action vectors vary in size across different tasks within a task set, necessitating specially designed input layers for each neural network to enable multi-task learning. We adopt the input-layer design from ODIS, as detailed in Appendix C of \cite{DBLP:conf/iclr/ZhangJLY0Z23}.}, thus it can potentially be used in a relevant but new task without retraining. However, generalization to a more complex task (e.g., 10m) could be challenging and fine-tuning the grouper with task-specific rewards may improve the overall learning performance. Yet, the fine-tuned case `grp-tuned' doesn't bring performance improvement, likely because the training of the grouper $h_\psi$ and high-level policy $\pi_\omega$ are interleaved and a carefully-designed co-training scheme is required. Last, if we change the source tasks for skill discovery from [3m, 5m] to [3m, 10m], the performance can be greatly boosted, as evidenced by `data-10m', showing the capability of VO-MASD-3D to extract effective skills from demonstrated data.

\section{Discussion of Limitations} \label{disc}

Regarding the limitations of this work, as discussed in Section \ref{utility}, the algorithm design of VO-MASD-3D could be further optimized for tasks involving heterogeneous agents. As a potential future direction, a co-training scheme for the high-level policy and the grouper in online MARL could be developed, enabling the grouping function to be fine-tuned with task-specific rewards for enhanced performance. Additionally, the proposed algorithms have broad practical applications, such as skill and policy learning for connected vehicles \cite{DBLP:conf/nips/PengLHLZ21} and drone swarms \cite{alharbi2024reinforcement}. Exploring their implementation in real-world scenarios, potentially by incorporating domain-specific knowledge, would be an exciting avenue for future work.

\end{document}